%% file: main.tex
\documentclass[journal,twoside,web]{ieeecolor}
\usepackage{generic}

\usepackage{textcomp}

\input{Ancillary/packages}
\input{Ancillary/commands}
\input{Ancillary/definitions}

\newcommand{\gitrepo}{https://github.com/LAVA-LAB/FBA}

\newcommand{\change}[1]{{#1}} 

\begin{document}

\title{Correct-by-construction reach-avoid control of partially observable linear stochastic systems}

\author{Thom Badings, Hasan A. Poonawala, Marielle Stoelinga, Nils Jansen
\thanks{This work was supported by NWO via the grant NWA.1160.18.238 (PrimaVera), the Department of Mechanical Engineering at the University of Kentucky, and by the ERC Starting Grant 101077178 (DEUCE).}
}

\maketitle

\begin{abstract}
We study feedback controller synthesis for reach-avoid control of discrete-time, linear time-invariant (LTI) systems with Gaussian process and measurement noise. The problem is to compute a controller such that, with at least some required probability, the system reaches a desired goal state in finite time while avoiding unsafe states. Due to stochasticity and nonconvexity, this problem does not admit exact algorithmic or closed-form solutions in general. Our key contribution is a correct-by-construction controller synthesis scheme based on a finite-state abstraction of a Gaussian belief over the unmeasured state, obtained using a Kalman filter. We formalize this abstraction as a Markov decision process (MDP). To be robust against numerical imprecision in approximating transition probabilities, we use MDPs with intervals of transition probabilities. By construction, any policy on the abstraction can be refined into a piecewise linear feedback controller for the LTI system. We prove that the closed-loop LTI system under this controller satisfies the reach-avoid problem with at least the required probability. The numerical experiments show that our method is able to solve reach-avoid problems for systems with up to 6D state spaces, and with control input constraints that cannot be handled by methods such as the rapidly-exploring random belief trees (RRBT).
\end{abstract}

\begin{IEEEkeywords}
Linear stochastic systems, nonlinear feedback control, formal abstraction, Markov decision process, Kalman filtering, partial observability.
\end{IEEEkeywords}

\input{1-Introduction}
\input{2-Preliminaries}
\input{3-Filtering}
\input{4-Abstraction}
\input{5-Correctness}
\input{6-Extensions}
\input{7-NumericalStudy}
\input{8-Conclusions}

\bibliographystyle{ieeetr}
\bibliography{references.bib}

\newpage

\begin{IEEEbiography}[{\includegraphics[width=1in,height=1.25in,clip,keepaspectratio]{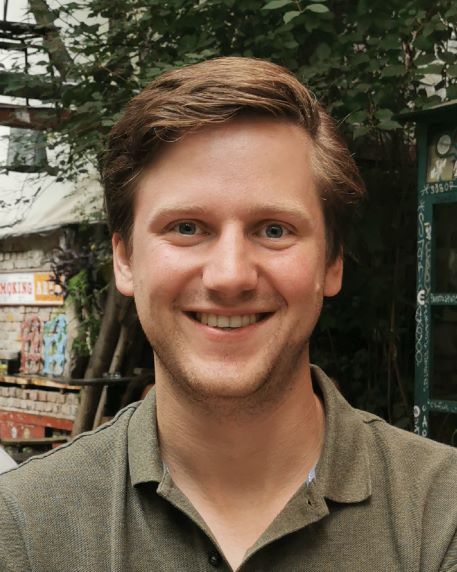}}]{Thom Badings} is a PhD candidate at the Institute for Computing and Information Science (iCIS) at the Radboud University, Nijmegen, The Netherlands.
He holds a B.Sc.\ (2017) and M.Sc.\ (2019, cum laude) degree in Industrial Engineering and Management from the University of Groningen.
His main research interests are on the intersection between control theory and formal methods.
Currently, he works on safe and robust sequential decision-making under uncertainty, with applications to autonomous and robotic systems, predictive maintenance, and power systems.

\end{IEEEbiography}

\begin{IEEEbiography}[{\includegraphics[width=1in,height=1.25in,clip,keepaspectratio]{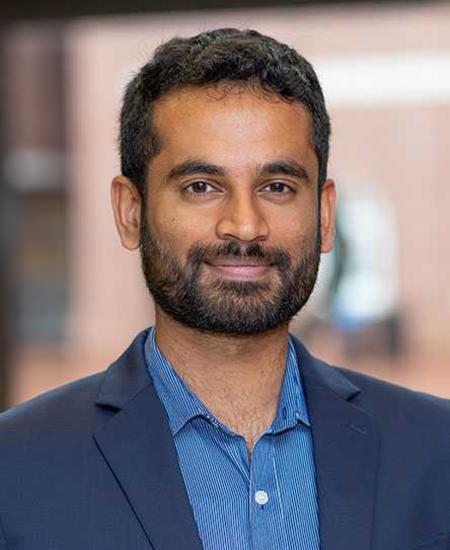}}]{Hasan A. Poonawala} is an assistant professor in the Department of Mechanical Engineering at the University of Kentucky. He holds a Master's degree in Mechanical Engineering from the University of Michigan (2009), and a Ph.D.\ in Electrical Engineering from the University of Texas at Dallas (2014). Dr.\ Poonawala worked as a postdoctoral researcher at the University of Texas at Austin, on combining AI and control theory. His research expertise spans mechatronics, control of multi-agent systems, vision-based motion control, and classifier-in-the-loop systems. His current research focuses on controlling robotic systems using high-dimensional sensor data, machine learning, and control theory.
\end{IEEEbiography}

\textbf{\begin{IEEEbiography}[{\includegraphics[width=1in,height=1.25in,clip,keepaspectratio]{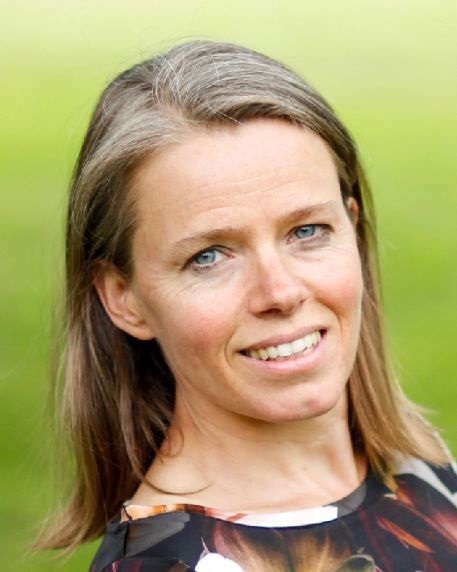}}]{Marielle Stoelinga} is a professor of risk management, working at the University of Twente, and the Radboud University Nijmegen, the Netherlands.
She holds a M.Sc.\ and a Ph.D.\ degree from the Radboud University Nijmegen, and has spent several years as a post-doc at the University of California at Santa Cruz, USA.
Prof Stoelinga holds several prestigious grants, including an ERC consolidator and a Dutch National Science Agenda grant,  funding the largest project on Predictive Maintenance in the Netherlands.
\end{IEEEbiography}}

\begin{IEEEbiography}[{\includegraphics[width=1in,height=1.25in,clip,keepaspectratio]{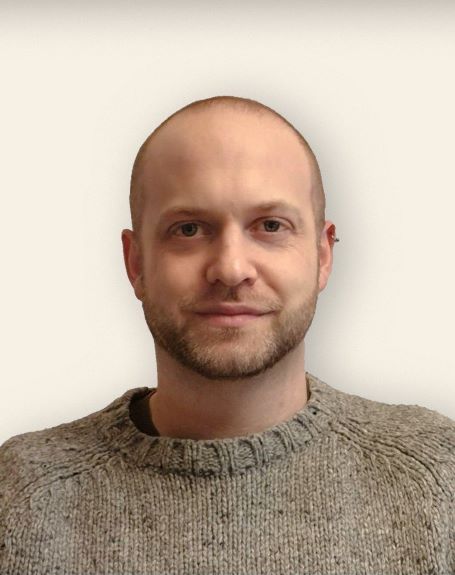}}]{Nils Jansen} is a tenured assistant professor at the Institute for Computing and Information Science (iCIS) at the Radboud University, Nijmegen, The Netherlands. He received his Ph.D. with distinction from RWTH Aachen University, Germany in 2015. Prior to Radboud University, he was a postdoc and research associate at the University of Texas at Austin. Dr. Jansen's current research is on formal reasoning about safety and dependability aspects in artificial intelligence (AI). At the heart of his research is the development of concepts from formal methods and control theory to reason about uncertainty and partial information in AI systems. He holds several grants within this area, both in academic and industrial settings. Dr. Jansen is a member of the European Lab for Learning and Intelligent Systems (ELLIS).
\end{IEEEbiography}

\end{document}

%% file: Ancillary/packages.tex
\usepackage{amsmath,amssymb,amsfonts}       
\usepackage{cases}
\usepackage{empheq}
\usepackage{mathbbol}
\usepackage{bm}
\usepackage{bbm}
\DeclareSymbolFontAlphabet{\amsmathbb}{AMSb}
\usepackage{mathtools}

\usepackage{amsthm}

\usepackage{siunitx}

\usepackage{adjustbox}

\DeclareMathOperator*{\minimize}{minimize}
\DeclareMathOperator*{\argmax}{arg\,max}

\usepackage[acronym]{glossaries}

\usepackage{fontawesome5}

\usepackage{graphicx}
\usepackage{xspace}
\usepackage{todonotes}
\usepackage{colonequals}
\usepackage[noadjust,nospace]{cite}
\usetikzlibrary{patterns,arrows,arrows.meta,calc,shapes,shadows,decorations.pathmorphing,decorations.pathreplacing,automata,shapes.multipart,positioning,shapes.geometric,fit,circuits,trees,shapes.gates.logic.US,fit, shadows.blur,shapes.symbols}

\usetikzlibrary{backgrounds,scopes}
\usepackage{marvosym}
\usepackage{multirow}
\usepackage{multicol}

\usepackage{tabularx}
\usepackage{booktabs}

\usepackage{caption}
\usepackage{subcaption}
\usepackage{url}
\usepackage[pdfpagelabels=true]{hyperref}
\usepackage{cleveref}

\crefname{equation}{Eq.}{Eq.}
\crefname{pluralequation}{Eqs.}{Eqs.}

\crefname{algorithm}{Algorithm}{Algorithm}
\crefname{pluralalgorithm}{Algorithms}{Algorithms}

\crefname{figure}{Fig.}{Fig.}
\crefname{pluralfigure}{Figs.}{Figs.}

\crefname{section}{Sect.}{Sect.}
\crefname{pluralsection}{Sects.}{Sects.}

\crefname{definition}{Def.}{Def.}
\crefname{pluraldefinition}{Defs.}{Defs.}

\crefname{theorem}{Theorem}{Theorem}
\crefname{pluraltheorem}{Theorems}{Theorems}

\crefname{lemma}{Lemma}{Lemma}
\crefname{plurallemma}{Lemmas}{Lemmas}

\crefname{example}{Example}{Example}
\crefname{pluralexample}{Examples}{Examples}

\crefname{assumption}{Assumption}{Assumption}
\crefname{pluralassumption}{Assumptions}{Assumptions}

\crefname{remark}{Remark}{Remark}
\crefname{pluralremark}{Remarks}{Remarks}

\crefname{table}{Table}{Table}
\crefname{pluraltable}{Tables}{Tables}

\crefname{problem}{Problem}{Problem}

\usepackage{tikz} 
\usepackage{pgfplots}
\usepgfplotslibrary{external}
\pgfplotsset{compat=1.8}

\definecolor{red}{rgb}{0.745,0.192,0.102}
\definecolor{darkgreen}{RGB}{34,161,55}
\definecolor{ruhuisstijlrood}{rgb}{0.745,0.192,0.102}
\definecolor{ruhuisstijlzwart}{rgb}{0,0,0}
\definecolor{ruhuisstijlwit}{rgb}{0.98,0.98,0.98}

\usepackage{mdframed}

\usetikzlibrary{arrows,decorations.pathmorphing,positioning,fit,trees,shapes,shadows,automata,calc} 

\usepackage{algorithm}
\usepackage{algorithmic}

\usepackage{changepage}

%% file: Ancillary/commands.tex
\newcommand{\ie}{i.e.\@\xspace}

\newcommand{\eg}{e.g.\@\xspace}

\newcommand{\prism}{\textrm{PRISM}\xspace}

\newcommand{\R}{\amsmathbb{R}}

\newcommand{\N}{\amsmathbb{N}}
\newcommand{\distr}[1]{\ensuremath{\mathit{Dist(#1)}}}

\newcommand{\States}{S}
\newcommand{\goalStates}{\ensuremath{\States_{G}}}
\newcommand{\criticalStates}{\ensuremath{\States_{C}}}
\newcommand{\Actions}{A}
\newcommand{\initState}{\ensuremath{s_I}}
\newcommand{\transfunc}{P}
\newcommand{\transfuncImdp}{\ensuremath{\mathcal{P}}}

\newcommand{\Interval}{\ensuremath{\amsmathbb{I}}}
\newcommand{\MDP}{\ensuremath{(\States,\Actions,\initState,\transfunc)}}
\newcommand{\iMDP}{\ensuremath{(\States,\Actions,\initState,\transfuncImdp)}}
\newcommand{\mdp}{\ensuremath{\mathcal{M}}}
\newcommand{\imdp}{\ensuremath{\mathcal{M}_\Interval}}

\newcommand{\policy}{\ensuremath{\pi}}
\newcommand{\policySpace}{\ensuremath{\Pi}}

\newcommand{\controller}{\ensuremath{\phi}}

\newcommand{\cStateSpace}{\ensuremath{\mathcal{X}}}
\newcommand{\cControlSpace}{\ensuremath{\mathcal{U}}}

\newcommand{\BeliefSpace}{\ensuremath{\mathcal{H}}}

\newcommand{\diag}[1]{\ensuremath{{\mathrm{diag}(#1)}}}

\newcommand{\Partition}{\ensuremath{\mathcal{R}}}
\newcommand{\goalRegion}{\ensuremath{\mathcal{X}_G}}
\newcommand{\criticalRegion}{\ensuremath{\mathcal{X}_C}}
\newcommand{\finiteHorizon}{\ensuremath{N}}

\newcommand{\belief}{\ensuremath{bel}}

\newcommand{\pproperty}{\ensuremath{\varphi_{\bm{x}_0}}}
\newcommand{\propertyAugm}{\ensuremath{\tilde\varphi_{\bm{\mean}_0}}}
\newcommand{\propertyMDP}{\ensuremath{\varphi_{\initState}}}
\newcommand{\propertyAugmMDP}{\ensuremath{\tilde\varphi_{\initState}}}

\newcommand{\Gauss[2]}{\ensuremath{\mathcal{N}(#1 , #2)}}
\newcommand{\mean}{\ensuremath{\mu}}
\newcommand{\cov}{\ensuremath{\Sigma}}

%% file: Ancillary/definitions.tex
\newtheorem{assumption}{Assumption}{\bfseries}{\itshape}
\newtheorem{theorem}{Theorem}{\bfseries}{\itshape}
{\bfseries}{\itshape}
\newtheorem{definition}{Definition}{\bfseries}{\itshape}
{\bfseries}{\itshape}
\newtheorem{remark}{Remark}{\bfseries}{\itshape}
\newtheorem{lemma}{Lemma}{\bfseries}{\itshape}
\newtheorem{problem}{Problem}{\bfseries}{\itshape}

\newacronym[]{LTI}{LTI}{linear time-invariant}
\newacronym[]{UAV}{UAV}{unmanned aerial vehicle}
\newacronym[plural=MDPs,firstplural=Markov decision processes (MDPs)]{MDP}{MDP}{Markov decision process}
\newacronym[plural=iMDPs,firstplural=interval MDPs (iMDPs)]{iMDP}{iMDP}{interval MDP}
\newacronym[plural=aMDPs,firstplural=augmented MDPs (aMDPs)]{aMDP}{aMDP}{augmented MDP}
\newacronym[plural=POMDPs,firstplural=partially observable MDPs (POMDPs)]{POMDP}{POMDP}{partially observable MDP}

\newacronym[]{RRT}{RRT}{rapidly-exploring random trees}
\newacronym[]{PRM}{PRM}{probabilistic road maps}

\newacronym[]{LQG}{LQG}{linear-quadratic-Gaussian}

\newacronym[]{PCTL}{PCTL}{probabilistic computation tree logic}
\newacronym[]{LTL}{LTL}{linear temporal logic}

%% file: 1-Introduction.tex
\section{Introduction}
\label{sec:Introduction}

Controlled autonomous systems are increasingly deployed in safety-critical settings~\cite{DBLP:journals/tiv/PadenCYYF16}.
Such systems are naturally modeled as partially observable stochastic systems:
partial observability models limited observability of state variables, whereas stochasticity accounts for factors of randomness and sensor imprecision.
A common task is to reach a desired goal region within a given time horizon while always avoiding collision with certain obstacles~\cite{kogel2023}, which is also  
called a \emph{reach-avoid property}~\cite{DBLP:conf/hybrid/SummersKLT11,DBLP:conf/hybrid/FisacCTS15,DBLP:conf/cdc/EsfahaniCL11}.
Reach-avoid tasks are ubiquitous, e.g., in motion planning (an
\gls{UAV} delivering a package while not crashing into buildings~\cite{DBLP:conf/cdc/HerbertCHBFT17}) and process control (increasing the level in a water tank without exceeding a threshold level~\cite{DBLP:journals/tac/YordanovTCBB12}).
The problem is to compute a controller such that the reach-avoid property is satisfied \emph{with at least some required probability}.
However, reach-avoid properties generally result in \emph{non-convex} constraints on the state set, which, together with input constraints, imply the need for \emph{nonlinear control laws}.
Moreover, to guarantee property satisfaction, we require algorithmic methods that reason explicitly over stochasticity.
While approaches such as \gls{LQG} control~\cite{anderson2007optimal,Datta2004}, Lyapunov methods~\cite{dequeiroz2000lyapunov}, and optimal control~\cite{fleming2012deterministic} reason about the stability and (asymptotic) convergence of systems, these methods can generally not satisfy these more challenging control requirements~\cite{belta2017formal}.

In this paper, we take a different perspective and leverage techniques from formal verification~\cite{DBLP:books/daglib/BaierKatoen2008} to compute feedback controllers that \emph{provably satisfy} a given reach-avoid property.
We consider discrete-time, \gls{LTI} systems with input constraints, and additive Gaussian \emph{process} and \emph{measurement noise} affecting the state transition and measurement model, respectively~\cite{Kulakowski2014dynamic,aastrom2012introduction,levine2018control,friedland2012control}.
The state variables represent the system's true state, but only measurements of the state are observed, thus capturing partial/limited observability of state variables.
Specifically, we consider the following problem:
\vspace{0.4em}
\begin{adjustwidth}{18pt}{18pt}
    Given a discrete-time \gls{LTI} system with additive Gaussian noise, a reach-avoid property, and a threshold probability, compute a feedback controller such that the induced closed-loop system satisfies this property with at least the given threshold probability.
\end{adjustwidth}
\vspace{0.4em}
To exemplify the hardness of the problem, consider a reach-avoid problem for a \gls{UAV} in stochastic wind conditions and with noisy sensors.
The problem is to compute a feedback controller that guarantees the property is satisfied with, \eg, a probability of at least $0.9$.
Property satisfaction is evaluated over state trajectories, so we must compute a controller while at the same time reasoning about the probability space induced by the closed-loop system under that controller.
\change{
Moreover, the controller must be bounded to ensure that the input constraints on the \gls{LTI} system are satisfied.
This problem does not admit an exact algorithmic or closed-form solution~\cite{DBLP:journals/automatica/BlondelT00}.
Sample-based methods generally cannot provide formal guarantees on property satisfaction under state and input constraints (as we show experimentally in \cref{subsec:experiments:UAV}),
whereas the efficiency of optimization-based and Lyapunov methods relies on the convexity of the problem, which is not the case here (see the related work and the references therein for details).
}

\begin{figure*}[t!]
	\centering
        \include{Includes/Fig1-Overview}
        \vspace{-2em}
	\caption{
        Layers of abstraction in our controller synthesis scheme.
        Given an \gls{LTI} system $\mathcal{S}$ and reach-avoid property $\pproperty$, we first apply Kalman filtering to obtain a dynamical system $\mathcal{B}$ for the mean $\bm\mean_k$ of the belief (ellipses depict covariances).
        We account for uncertainty between the mean $\bm\mean_k$ and (unobservable) state $\bm{x}_k$ by expanding (contracting) critical (goal) regions by a time-varying error bound $\varepsilon_{k} \geq 0$. 
        Second, we abstract the belief mean dynamics $\mathcal{B}$ with the expanded/contracted regions as interval MDP (iMDP) $\imdp$.
        We compute an optimal policy for this iMDP which we refine to a controller for the \gls{LTI} system.
        }
	\label{fig:Overview}
\end{figure*}

\subsection*{Our filter-based abstraction scheme}
Our main contribution is a correct-by-construction controller synthesis scheme to solve the problem above.
Our approach is shown in \cref{fig:Overview} and combines filtering to alleviate partial observability, with abstraction to solve the controller synthesis problem.
We explain the key steps of our approach.

\subsubsection*{Step 1 -- Filtering}
We employ a Kalman filter~\cite{Kalman1960,DBLP:books/daglib/Thrun2005} to represent the belief as a Gaussian distribution.
The update of the mean of this Gaussian belief ($\bm\mean_k$ in \cref{fig:Overview}) is defined as a \emph{fully observable} linear system with additive Gaussian process noise, which we call the \emph{belief (mean) system}.
On the other hand, the update of the belief covariance is deterministic and independent of the controller, enabling us to split the problem into two parts.
First, we account for the belief covariance (representing the uncertainty between the belief mean and the actual state) by defining an augmented property, whose sets of goal states (critical states) are suitably contracted (expanded) by a pre-computed and time-varying \emph{error bound}~\cite{DBLP:conf/cdc/HerbertCHBFT17, DBLP:conf/icra/Fridovich-KeilH18}.
By taking this error bound such that it upper bounds the distance between the mean of the belief and the actual state with high probability, we show that we can reduce the problem of controlling the \gls{LTI} system to a control problem on the mean of the belief with respect to the augmented property (\cref{thm:BeliefCorrectness}).

\subsubsection*{Step 2 -- Abstraction}
We abstract the \emph{dynamics for the mean of the belief} into a \gls{MDP}~\cite{DBLP:books/wi/Puterman94,DBLP:books/daglib/BaierKatoen2008}.
Abstract states correspond to a partition of the continuous state space, and actions capture control inputs that induce stochastic transitions between these states.
A key distinguishing feature of our abstraction is that we use \emph{backward reachability computations} on the belief dynamics to determine which abstract actions are enabled in each state, yielding an abstraction that is sound by construction.
By contrast, other abstractions (see the related work or, e.g., \cite{LSAZ21} for details) often rely on \emph{forward reachability computations} through a discretization of the control input space, leading to abstraction errors.
Our backward method avoids such errors at the cost of requiring slightly more restrictive assumptions on the \gls{LTI} system (see \cref{assump:PredSetRank}).
Computing the transition probabilities of the abstract \gls{MDP} involves integrating multivariate Gaussian distributions, which cannot be done exactly~\cite{Cunningham2011gaussian,Genz2000multivariateGaussians}.
\change{
Thus, we capture the inherent numerical imprecision in estimating these probabilities by using \emph{intervals of probabilities}, which we embed in a so-called \gls{iMDP}~\cite{DBLP:journals/ai/GivanLD00}, also known as robust MDP~\cite{DBLP:journals/mor/WiesemannKR13}.
}

\subsubsection*{Step 3 -- Controller synthesis}
With methods such as value iteration, we can efficiently compute policies that maximize the probability of satisfying reach-avoid properties~\cite{DBLP:books/wi/Puterman94}.
For \glspl{iMDP}, a policy has to robustly account for \emph{all possible probabilities} within the intervals~\cite{DBLP:conf/cav/PuggelliLSS13, DBLP:conf/cdc/WolffTM12}.
Such policies for \glspl{iMDP} can be computed using robust versions of value iterations, which are implemented in, \eg, the probabilistic model checker \prism~\cite{DBLP:conf/cav/KwiatkowskaNP11}.
We show that any policy on the \gls{iMDP} can be refined into a \emph{piecewise linear feedback controller} for the \gls{LTI} system.
Crucially, the probability of satisfying the reach-avoid property on the \gls{iMDP} is a lower bound on the satisfaction probability for the \gls{LTI} system (\cref{thm:Correctness,thm:correctness:2phase}).

\subsubsection*{Two-phase time horizon}
The size of the abstract \gls{iMDP} grows with the granularity of the state space partitioning and the (finite) time horizon of the property.
To reduce the computational complexity, we 
divide the finite horizon of $N \in \N$ steps into two phases: 1) a \emph{transient phase} of time steps $0,\ldots,\bar{N}-1$ in which every time step is modeled explicitly, and 2) a \emph{steady-state phase} which lumps steps $\bar{N},\ldots,N$ into a single step.
The relative length of these two phases provides a trade-off between the size of the \gls{iMDP}, versus the level of conservatism of the obtained performance guarantees.

\subsection*{Related work}
\label{sec:related_work}
We give an overview of approaches to solving (reach-avoid) control problems for partially observable stochastic systems.

Using (i)\gls{MDP} abstractions for verification and controller synthesis has been widely studied in general~\cite{Abate2008probabilisticSystems, Alur2000, LSAZ21, DBLP:journals/siamads/SoudjaniA13, DBLP:journals/tac/LahijanianAB15, Huijgevoort2023SySCoRe}.
Under a (bi)simulation relation~\cite{hermanns2011probabilistic,DBLP:conf/cdc/SchonHHS22}, policies on the abstract model can be refined~\cite{DBLP:journals/tac/ReissigWR17} to controllers for the continuous system with formal guarantees.
In control, discrete abstractions are commonly called \emph{symbolic models}~\cite{DBLP:journals/tac/GirardPT10,Pola2015,DBLP:journals/tac/SinyakovG23}.
Our abstraction scheme is most similar to~\cite{DBLP:conf/aaai/BadingsA00PS22, Badings2022JAIR, Badings2023AAAI, DBLP:journals/corr/Rickard}; however, these papers (and in fact, most abstraction schemes in general~\cite{LSAZ21}) consider fully observable systems.
Existing abstractions for partially observable systems use (computationally expensive) \glspl{POMDP}~\cite{DBLP:conf/hybrid/LesserO15, DBLP:journals/tac/LesserO17} or assume maximum likelihood observations~\cite{DBLP:conf/rss/PlattTKL10} or a constant belief covariance~\cite{DBLP:conf/adhs/HaesaertNVTAAM18}.
By contrast, we avoid such assumptions by abstracting the stochastic dynamics for the mean of the belief directly, whereas we account for the uncertainty between the mean and the actual state by augmenting the property.

Sample-based algorithms such as \gls{RRT}~\cite{DBLP:journals/ijrr/LaValleK01} and \gls{RRT}*~\cite{DBLP:conf/rss/KaramanF10} efficiently find paths through (partially) known environments.
\change{Extensions of sample-based algorithms to partially observable systems include \gls{RRT} in belief space (RRBT)~\cite{DBLP:conf/icra/BryR11}, SLAP~\cite{DBLP:journals/trob/Agha-mohammadiA18}, the belief roadmap~\cite{DBLP:conf/isrr/PrenticeR07}, and algorithms based on Monte Carlo tree search~\cite{DBLP:journals/trob/SunSPK21}.
These methods have been effective in practice but only converge in the limit of infinite samples~\cite{DBLP:journals/ijrr/KaramanF11}.
Moreover, methods such as SLAP rely on point estimates that are inherently associated with \emph{statistical errors} and cannot give \emph{formal guarantees} on satisfying reach-avoid problems, as we do in this paper.
While RRBT does provide formal guarantees, the method relies on linear tracking controllers that require the input space to be unbounded.
In the experiments in \cref{sec:NumericalStudy}, we show that RRBT uses incorrect uncertainty predictions if the input space is bounded, whereas our method yields correct-by-construction controllers under any convex constraints on the input space.}

Control barrier functions (CBFs) certify liveness and safety of (stochastic) dynamical systems~\cite{clark2021control,DBLP:journals/tac/PrajnaJP07,DBLP:journals/tac/AmesXGT17,DBLP:journals/tac/JagtapSZ21,DBLP:journals/csysl/LindemannD19}.
Some examples include temporal logic verification~\cite{DBLP:conf/atva/JagtapSZ18,DBLP:journals/tac/AhmadiJWT21}
and controller synthesis~\cite{DBLP:journals/csysl/JahanshahiJZ21,DBLP:journals/tcns/JahanshahiLZ23} of partially observable stochastic systems.
However, depending on the property to be verified and the convexity of the problem, finding CBFs can be challenging.

Hamilton-Jacobi reachability analysis~\cite{DBLP:conf/cdc/BansalCHT17} is used for planning under full observability in FaSTrack~\cite{DBLP:conf/cdc/HerbertCHBFT17}.
Various papers develop hierarchical approaches to reach-avoid control for nonstochastic linear systems~\cite{kogel2023,DBLP:journals/tac/FanQMNMV22,Bofan2023}.
Generalizations of \gls{LQG}~\cite{anderson2007optimal} to problems with obstacles also exist~\cite{DBLP:conf/icra/BergWGNM12,DBLP:journals/tase/SunBA16}; however these methods consider cost minimization and cannot provide formal guarantees on temporal (e.g., reach-avoid) properties.
Optimization-based approaches also exist, e.g., based on model predictive control~\cite{Rosolia2020unified,Maia2009}, and tools such as SReachTools~\cite{DBLP:conf/hybrid/VinodGO19}.
Reach-avoid verification based on convex optimization for continuous-time but nonstochastic systems was recently considered by~\cite{DBLP:journals/corr/abs-2208-08105}.
However, for general reach-avoid properties (having non-convex state constraints), the resulting optimization problems are non-convex.

\subsection*{Paper outline}
We formalize the problem in \cref{sec:Preliminaries}.
In \cref{sec:BeliefSpacePlanning}, we use Kalman filtering to define the belief evolution as a stochastic dynamical system, and we introduce our \gls{iMDP} abstraction procedure in \cref{sec:FilterBasedAbstraction}.
We present our algorithm and its correctness in \cref{sec:Algorithm}, and the two-phase time horizon extension in \cref{sec:2phasehorizon}.
Finally, we present our experiments in \cref{sec:NumericalStudy}.

%% file: Includes/Fig1-Overview.tex
\newcommand\goal{(3.7cm, 3.3cm)}
\newcommand\goalsizeX{1.6cm}
\newcommand\goalsizeY{1.4cm}

\newcommand\goalsizeXaug{1.0cm}
\newcommand\goalsizeYaug{0.7cm}

\newcommand\crit{(3.5cm, 1.3cm)}
\newcommand\critsizeX{1cm}
\newcommand\critsizeY{1cm}

\newcommand\critsizeXaug{2.0cm}
\newcommand\critsizeYaug{2.0cm}

\newcommand\tikzX{(0.6cm, 0.5cm)}
\newcommand\tikzXnext{(1.5cm, 2.7cm)}

\newcommand\tikzMean{(0.6cm, 1.0cm)}
\newcommand\tikzMeanPred{(1.2cm, 2.8cm)}
\newcommand\tikzMeanNext{(1.7cm, 2.3cm)}

\newcommand\tikzYnext{2.3cm}

\resizebox{0.93\textwidth}{!}{%
\begin{tikzpicture}[scale=1.0]
    \node[] at (2.4cm, 4.3cm) [above, align=center] 
    {\begin{tabular}{ll} \toprule
        \textbf{LTI system} & $\mathcal{S}$ \\
        \midrule
        State variables & $\bm{x}_k \in \R^n$ \\
        Measurement & $\bm{y}_k \in \R^q$ \\
        Reach-avoid property & $\pproperty$ \\
        \bottomrule
        \end{tabular}};

    \draw [->] (0,0) -- (4.8cm,0cm) node [midway, below] {$\bm{x}^{(1)}$};
    \draw [->] (0,0) -- (0cm,4cm) node [at end, left] {$\bm{x}^{(2)}$};

    \node[] (pred_mean) at \tikzXnext {};     
    \node[] (y_next) at (0, \tikzYnext) {};
    
    \node[] (mean) at \tikzX [circle,fill=red,inner sep=1pt] {};
    \node[right] at (mean) {$\bm{{x}}_k$};
    \node[] at (pred_mean) [circle,fill=red,inner sep=1pt] {};
    \node[above] at (pred_mean) {$\bm{{x}}_{k+1}$};
    
    \draw [->] (mean) to [out=90,in=-135] node[right, pos=0.6]{$\bm{u}_k$} (pred_mean);
    
    \node[] (crit) at \crit {};
    \node[] (goal) at \goal {};
    
    \node(c1)[rectangle, 
            draw = red,
            minimum width = \critsizeX,
            minimum height = \critsizeY,
            fill = red!50] at (crit) {$\goalRegion$};
            
    \node(g1)[rectangle, 
            draw = OliveGreen,
            minimum width = \goalsizeX,
            minimum height = \goalsizeY,
            fill = OliveGreen!30] at (goal) {$\goalRegion$};

    \draw [densely dashed, gray] (0, \tikzYnext) -- (4.8, \tikzYnext) {};
    \node[] at (y_next) [diamond,fill=gray,inner sep=1.5pt] {};
    \node[left] at (y_next) {$\bm{y}_{k+1}$};   

    \draw[-{Triangle[width=18pt,length=8pt]}, line width=10pt] (5.5cm, 5cm) -- (6.1cm, 5cm) node [midway, above, yshift=0.1cm] {Filtering};
\end{tikzpicture} 
\hspace{-2.4em}
\begin{tikzpicture}[scale=1.0]
    \node[] at (2.4cm, 4.64cm) [above, align=center] 
    {\begin{tabular}{ll} \toprule
        \textbf{Belief mean system} & $\mathcal{B}$ \\
        \midrule
        State variables & $\bm\mean_k \in \R^n$ \\
        Augmented property & $\propertyAugm$ \\
        \bottomrule
        \end{tabular}};

    \draw [->] (0,0) -- (4.8cm,0cm) node [midway, below] {$\bm\mean^{(1)}$};
    \draw [->] (0,0) -- (0cm,4cm) node [at end, left] {$\bm\mean^{(2)}$};

    \node[] (mean) at \tikzMean {};
    \node[] (pred_mean) at \tikzMeanPred {};
    \node[] (mean_next) at \tikzMeanNext {};
    \node[] (y_next) at (0, \tikzYnext) {};
    
    \def\particles{(mean) }
        \foreach \point in \particles{
          \foreach\i in {0,0.08,...,0.8} {
            \fill[opacity=0.7*\i,BurntOrange,rotate around={20:\point}] \point ellipse ({.4-\i/2} and {0.75-1*\i});         
          }
        } 
        
    \def\particles{(pred_mean) }
        \foreach \point in \particles{
          \foreach\i in {0,0.08,...,0.8} {
            \fill[opacity=0.7*\i,BurntOrange,rotate around={30:\point}] \point ellipse ({.8-\i} and {1.5-2*\i});         
          }
        } 

    \def\particles{(mean_next) }
        \foreach \point in \particles{
          \foreach\i in {0,0.08,...,0.8} {
            \fill[opacity=0.7*\i,OliveGreen,rotate around={-30:\point}] \point ellipse ({(.8-\i)/1.5} and {(1.5-2*\i)/1.5});         
          }
        } 
    
    \node[] at \tikzMean [circle,fill=red,inner sep=1pt] {};
    \node[right] at (mean) {$\bm{\mean}_k$};
    
    \node[] at (pred_mean) [circle,fill=red,inner sep=1pt] {};
    \node[above] at (pred_mean) {$\hat{\bm{\mean}}_{k+1}$};

    \node[] at (mean_next) [circle,fill=red,inner sep=1pt] {};
    \node[below] at (mean_next) {$\bm{\mean}_{k+1}$};
    
    \node[] at (pred_mean) [black, anchor=south, xshift=0.2cm, yshift=0.8cm] {$\cov_{\delta_{k+1}}$};

    \node[] at (mean_next) [black, anchor=south, xshift=0.7cm, yshift=0.4cm] {$\cov_{k+1}$};
    
    \draw [->] (mean) to [out=90,in=-135] node[right, pos=0.6]{$\bm{u}_k$} (pred_mean);

    \draw[->] (pred_mean) -- (mean_next) node {};
    
    \node[] (crit) at \crit {};
    \node[] (goal) at \goal {};
    
    \node(c1)[rectangle, 
            draw = red,
            minimum width = \critsizeX,
            minimum height = \critsizeY,
            fill = red!50] at (crit) {};
    \node[xshift=-0.5cm, yshift=-0.75cm] at (c1) {$\criticalRegion^{\varepsilon_{k+1}}$};
    \node(c2)[rectangle,
            draw = red,dashed,
            minimum width = \critsizeXaug,
            minimum height = \critsizeYaug] at (crit) {};
            
    \node(g1)[rectangle, 
            draw = OliveGreen,
            minimum width = \goalsizeX,
            minimum height = \goalsizeY,
            fill = OliveGreen!30] at (goal) {$\goalRegion^{\varepsilon_{k+1}}$};
    \node(g2)[rectangle,
            draw = OliveGreen,dashed,
            minimum width = \goalsizeXaug,
            minimum height = \goalsizeYaug] at (goal) {};    
    
    \draw[->] (c1.south) -- (c2.south) node [pos=0.5, right, align=center] {$\varepsilon_{k+1}$};
    
    \draw[->] (g1.north) -- (g2.north) node [pos=0.5, right, align=center] {$\varepsilon_{k+1}$};

    \draw [densely dashed, gray] (0, \tikzYnext) -- (mean_next) {};
    \node[] at (y_next) [diamond,fill=gray,inner sep=1.5pt] {};
    \node[left] at (y_next) {$\bm{y}_{k+1}$};    

    \draw[-{Triangle[width=18pt,length=8pt]}, line width=10pt] (5.7cm, 5cm) -- (6.3cm, 5cm) node [midway, above, yshift=0.1cm] {Abstraction};
\end{tikzpicture} 
\hspace{-1.2em}
\begin{tikzpicture}[scale=1.0,
                    nodestyle/.style={draw,circle},
                    ]

    \node[] at (2.4cm, 4.64cm) [above, align=center] 
    {\begin{tabular}{ll} \toprule
        \textbf{Abstract iMDP} & $\imdp$ \\
        \midrule
        State (discrete) & $s \in \States$ \\
        Augmented property & $\propertyAugmMDP$ \\
        \bottomrule
        \end{tabular}};

    \node[] (mean) at (0.4cm, 1.2cm) {};
    \node[] (pred_mean) at \tikzMeanPred {};
    \node[] (mean_next) at \tikzMeanNext {};
    \node[] (y_next) at (0, \tikzYnext) {};

    \node[] (crit) at \crit {};
    \node[] (goal) at \goal {};
    
    \newcommand\boxW{0.8}
    \newcommand\uppX{6}
    \newcommand\uppY{5}

    \foreach \x in {1,...,\uppX} 
    {
        \draw[gray!70, dashed] (\x*\boxW, 0) -- (\x*\boxW,\boxW*\uppY);
    }
    \foreach \y in {1,...,\uppY} 
    {
        \draw[gray!70, dashed] (0, \y*\boxW) -- (\boxW*\uppX, \y*\boxW);
    }
    \draw[gray!70, dashed] (3.2cm, 3.65cm) --(4.8cm, 3.65cm);
    \draw[gray!70, dashed] (3.2cm, 2.95cm) --(4.8cm, 2.95cm);
    \draw[gray!70, dashed] (4.2cm, 4.00cm) --(4.2cm, 2.40cm);

    \draw[gray!70, dashed] (2.5cm, 0cm) --(2.5cm, 2.4cm);
    \draw[gray!70, dashed] (4.5cm, 0cm) --(4.5cm, 2.4cm);
    \draw[gray!70, dashed] (2.4cm, 2.3cm) --(4.8cm, 2.3cm);
    \draw[gray!70, dashed] (2.4cm, 0.3cm) --(4.8cm, 0.3cm);
    
    \node[] at (mean) [] {$s$};
    \node[] at (pred_mean) [] {$s'$};
    \node[] (successorA) at (2.0cm, 2.8cm) {};
    \node[] at (successorA) [] {$s''$};
    \node[] (successorB) at (2.0cm, 2.0cm) {};
    \node[] at (successorB) {$s'''$};

    \node(c2)[rectangle,
            fill = red!20,
            minimum width = \critsizeXaug,
            minimum height = \critsizeYaug] at (crit) {};
    \node (c2_label) at (c2) [xshift=0.2cm] {$s_\star$};
            
    \node(g2)[rectangle,
            fill = OliveGreen!20,
            minimum width = \goalsizeXaug,
            minimum height = \goalsizeYaug] at (goal) {$s_g$};    

    \draw [->] (mean) to [out=20,in=-60] node[left, pos=0.7, xshift=0.07cm]{$[0.4, 0.6]$} (pred_mean);
    \draw [->] (mean) to [out=20,in=-110] node[right, pos=1.0]{$[0.2, 0.35]$} ($(successorA) + (0,-0.2cm)$);
    \draw [->] (mean) to [out=20,in=-100] node[right, pos=0.75, xshift=0.1cm]{$[0.2, 0.4]$} ($(successorB) + (0,-0.2cm)$);
    \draw [->] (mean) to [out=20,in=-135] node[below, pos=0.1]{$a$} node[below, pos=0.6]{$[0.05, 0.1]$} (c2_label);

    \draw [gray!70, dashed] (0,0) -- (4.8cm,0cm) node [midway, below, white] {$\bm\mean^{(1)}$};
    \draw [gray!70, dashed] (0,0) -- (0cm,4cm) node [] {};

    \draw[-{Triangle[width=18pt,length=8pt]}, line width=10pt] (5.4cm, 5cm) -- (6.0cm, 5cm) node [midway, above, yshift=0.1cm, align=center] {Controller \\ refinement};
\end{tikzpicture} 
}

%% file: 2-Preliminaries.tex
\section{Foundations and outline}
\label{sec:Preliminaries}
We denote by $\N_0 = \{0\} \cup \N$ the set of natural numbers including zero. 
A \emph{discrete probability distribution} over a finite set $X$ is a function $\mathit{prob} \colon X \to [0,1]$ with $\sum_{x \in X} \mathit{prob}(x) = 1$.
The set of all distributions over $X$ is $\distr{X}$, and the number of elements in a set $X$ is $|X|$.
All vectors $\bm{x} \in \R^n$, with $n \in \N$, are denoted by bold letters and are column vectors.
Moreover, $I_n$ denotes the $n \times n$ identity matrix, $\bm{x}_{1:n}$ denotes a vector $[\bm{x}_1^\top, \ldots, \bm{x}_n^\top]^\top$, and for a vector $\bm{x} \in \R^n$, $\diag{\bm{x}}$ is the square matrix with the $\bm{x}$ its diagonal and $0$ elsewhere.
A multivariate Gaussian random variable $\bm{z} \sim \Gauss[\bm\mean]{\cov} \in \R^n$ is defined by its mean vector $\bm{\mean} \in \R^n$ and positive semi-definite covariance matrix $\cov \in \R^{n\times n}$~\cite{tong2012multivariate}.

\subsection{LTI systems}
Consider a discrete-time \gls{LTI} system $\mathcal{S}$, whose continuous state $\bm{x}_{k} \in \R^n$ evolves over discrete time steps $k \in \N_0$ as%
\begin{subequations}%
    \begin{empheq}[left=\mathcal{S} \colon \empheqlbrace]{align}%
        \, \bm{x}_{k+1} &= A \bm{x}_k + B \bm{u}_k + \bm{w}_k,
        \quad \bm{x}_0 \in \R^n
		\label{eq:LTI_system_process}
        \\
        \, \bm{y}_{k+1} &= C \bm{x}_{k+1} + \bm{v}_{k+1},
		\label{eq:LTI_system_measurement}
    \end{empheq}%
    \label{eq:LTI_system}%
\end{subequations}%
\noindent
where $\bm{y}_{k} \in \R^q$ is the measurement of the state, $\bm{u}_k \in \cControlSpace$ is the control input, constrained by a bounded convex set $\cControlSpace \subset \R^p$, and $\bm{w}_k \sim \Gauss[\bm\mean_{w_k}]{\cov_{w_k}}$ and $\bm{v}_k \sim \Gauss[0]{\cov_{v_k}}$ are Gaussian process and measurement noise terms, respectively (which model imperfect actuation and sensing).
The state $\bm{x}_{k+1}$ is a linear function of the state and control input at time $k$ via the \emph{system matrix} $A \in \R^{n \times n}$ and the \emph{input matrix} $B \in \R^{n \times p}$.
Similarly, the measurement is a linear function of the state through the \emph{observation matrix} $C \in \R^{q \times n}$.
Due to linearity and Gaussian noise, system $\mathcal{S}$ is commonly called a \emph{linear Gaussian system}.
If matrix $C$ is not invertible, the measurements are \emph{limited}, since the state cannot be reconstructed from a single measurement.

\subsubsection{Belief distribution}
The measurement noise and possibly limited measurements in the \gls{LTI} system result in imprecise knowledge of the actual state $\bm{x}_k$ at any time $k$. %
We define the available knowledge of this state by a \emph{belief distribution}.
The belief $\belief(\bm{x}_k) \in \distr{\R^n}$ over a state $\bm{x}_k$ at time $k$ is as defined as follows~\cite{DBLP:books/daglib/Thrun2005}:
\begin{definition}	
\label{def:Belief}
A belief $\belief(\bm{x}_k) \in \distr{\R^n}$ over $\bm{x}_k$ is defined by the posterior distribution $\belief(\bm{x}_k) = p(\bm{x}_k \, | \, \bm{y}_{1:k}, \bm{u}_{1:k})$, with $\bm{y}_{1:k}$ and $\bm{u}_{1:k}$ all measurements and inputs up to time $k$.
\end{definition}%

\subsubsection{Controller}
We consider time-varying feedback controllers for the \gls{LTI} system in \cref{eq:LTI_system} of the following form:
\begin{definition}
    \label{def:controller}
    A time-varying feedback controller is a function $\controller \colon \BeliefSpace \times \N_0 \to \cControlSpace$, which maps a belief $\belief(\bm{x}_k) \in \BeliefSpace$ of the state $\bm{x}_k$ and a time step $k \in \N_0$ to a control input $\bm{u}_k \in \cControlSpace$.
\end{definition}
For brevity, we denote by $\mathcal{S}(\controller)$ the closed-loop version of system $\mathcal{S}$ in which, for each $k \in \N_0$, the input $\bm{u}_k \coloneqq \controller(\belief(\bm{x}_k), k)$ is determined by the controller $\controller$.

\subsubsection{Reach-avoid properties}
We consider control problems formalized as reach-avoid properties:
\begin{definition}
    \label{def:ReachAvoidProperty}
    A reach-avoid property is a tuple $\pproperty = (\goalRegion, \criticalRegion, \bm{x}_0, N)$, where $\goalRegion, \criticalRegion \subset \R^n$ are compact sets of goal and critical states, respectively, with $\goalRegion \cap \criticalRegion = \varnothing$, $\bm{x}_0 \in \R^n$ is an initial state, and $\finiteHorizon \in \N$ is a finite time horizon.
\end{definition}
Due to the noise terms $\bm{w}_k$ and $\bm{v}_k$, the trajectory generated by system $\mathcal{S}(\controller)$ is a realization from a stochastic process.
We say that a finite trajectory $\bm{x}_0, \bm{x}_1, \ldots, \bm{x}_N$ generated by system $\mathcal{S}(\controller)$ satisfies property $\pproperty$ if there exists a $k \leq N$ such that $\bm{x}_k \in \goalRegion$ (reaching a goal state), while $\bm{x}_{k'} \not\in \criticalRegion \forall k' \in \{0,\ldots,k\}$ (avoiding critical states until then).
The probability $\Pr(\mathcal{S}(\controller) \models \pproperty)$ that $\mathcal{S}(\controller)$ satisfies $\pproperty$ is defined as follows.
\begin{definition}
The satisfaction probability $\Pr(\mathcal{S}(\controller) \models \pproperty)$ that system $\mathcal{S}(\controller)$ generates a trajectory satisfying $\pproperty$ is
\begin{align}%
\Pr & (\mathcal{S}(\controller) \models \pproperty) = 
\amsmathbb{P} \Big\{
    \bm{x}_0, \bm{x}_1, \ldots, \bm{x}_N
    \, \colon \,
    \\
    & 
    \exists k \in \{0,\ldots,\finiteHorizon\},
    \, 
    \bm{x}_{k} \in \goalRegion,
    \,\,
    \bm{x}_{k'} \notin \criticalRegion \, \forall k' \in \{0,\ldots,k \}
    \Big\}.
    \nonumber
\end{align}
\end{definition}

\subsection{Formal problem statement}
\label{sec:FormalProblem}

In this paper, we develop a method to solve the problem stated in \cref{sec:Introduction}.
Formally, we solve the following problem:
\begin{problem}
\label{prob:Problem}
Given an \gls{LTI} system $\mathcal{S}$ defined by \cref{eq:LTI_system}, a reach-avoid property $\pproperty = (\goalRegion, \criticalRegion, \bm{x}_0, N)$, and a desired threshold probability $\eta \in [0,1]$, compute a controller $\controller$ such that $\Pr(\mathcal{S}(\controller) \models \pproperty) \geq \eta$.
\end{problem}
Observe that the problem is not to find an optimal controller, i.e., one that maximizes the probability $\Pr(\mathcal{S}(\controller) \models \pproperty)$, but instead one with a satisfaction probability above threshold $\eta$, which thus acts as a minimum required performance level.

\subsection{Markov decision processes}
\label{sec:Preliminaries_MDPs}

We now introduce the discrete-state models that we use to formalize the abstractions that we construct in this paper.

\begin{definition}
	A \glsfirst{MDP} is a tuple $\mdp=\MDP$ where $\States$ is a finite set of states, $\Actions$ is a finite set of actions, $\initState$ is the initial state, and $\transfunc \colon \States \times \Actions \rightharpoonup \distr{\States}$ is the (partial) probabilistic transition function.
	\label{def:MDP}
\end{definition}

We call a tuple $(s,a,s')$ with probability $\transfunc(s,a)(s')>0$ a \emph{transition}.
The nondeterministic choices of actions in an \gls{MDP} are resolved by \emph{policies}.
A deterministic (or pure) \emph{policy}~\cite{DBLP:books/daglib/BaierKatoen2008} for an \gls{MDP} is a function $\policy \colon \States^* \to \Actions$, where $\States^*$ is a sequence of states, and the set of all such policies for \gls{MDP} $\mdp$ is denoted by $\policySpace_\mdp$.
A reach-avoid property $\propertyMDP = (\goalStates, \criticalStates, \initState, N)$ for an \gls{MDP} is defined similarly to \cref{def:ReachAvoidProperty}, but with $\goalStates, \criticalStates \subset \States$ and $\initState \in \States$ defined with respect to the \emph{discrete} set of states of the \gls{MDP}.
An \gls{MDP} $\mdp$ with a policy $\policy$ induces a Markov chain, which we denote by $\mdp(\policy)$.
The (reach-avoid) probability of satisfying $\propertyMDP$ under policy $\policy$ is written as $\Pr(\mdp(\pi) \models \propertyMDP)$.
An optimal policy $\policy^* \in \policySpace_\mdp$ maximizes the reach-avoid probability:
\begin{equation}
    \policy^* = \argmax_{\policy \in \policySpace_\mdp} \Pr(\mdp(\pi) \models \propertyMDP).
	\label{eq:optimalPolicy}
\end{equation}
Deterministic policies suffice to obtain this optimum~\cite{DBLP:books/wi/Puterman94}.
Interval MDPs (iMDPs) extend \glspl{MDP} with \emph{intervals} of transition probabilities instead of concrete values:
\begin{definition}
	An \glsfirst{iMDP} is a tuple $\imdp=\iMDP$ where $\States$, $\Actions$, and $\initState$ are defined by \cref{def:MDP}, and where the uncertain (partial) probabilistic transition function $\transfuncImdp \colon \States \times \Actions \times \States \rightharpoonup \Interval \cup \{ [0,0] \}$ is defined over intervals $\Interval = \{ [a,b] \ | \ a,b \in (0,1] \text{ and } a \leq b \}$.
	\label{def:iMDP}
\end{definition}
An \gls{iMDP} defines a (possibly empty) set of \glspl{MDP} that vary only in their transition function.
For an \gls{MDP} with transition function $\transfunc \colon \States \times \Actions \rightharpoonup \distr{\States}$, we write $\transfunc \in \transfuncImdp$ if for all $s,s' \in \States$ and $a \in \Actions$ we have $\transfunc(s,a)(s') \in \transfuncImdp (s,a,s')$ and $P(s, a)\in\distr{\States}$.
In particular, for an \gls{MDP} with transition function $\transfunc \colon \States \times \Actions \rightharpoonup \distr{\States}$, we write $\transfunc \in \transfuncImdp$ if for all $s,s' \in \States$ and $a \in \Actions$ we have $\transfunc(s,a)(s') \in \transfuncImdp (s,a,s')$ and $P(s, a)\in\distr{\States}$.
The fact that an interval cannot have a zero lower bound except for the $[0,0]$ interval implies that the \emph{graph} of each \gls{MDP} $\transfunc \in \transfuncImdp$ is the same.
As is common for \glspl{iMDP}~\cite{DBLP:conf/cav/PuggelliLSS13,DBLP:conf/cdc/WolffTM12}, we consider policies with reach-avoid probabilities that are \emph{robust} against any choice of probabilities $\transfunc \in \transfuncImdp$.
Specifically, we compute an optimal policy $\policy^* \in \policySpace_{\imdp}$ for \gls{iMDP} $\imdp$ that maximizes the lower bound on the reach-avoid probability over all $\transfunc \in \transfuncImdp$:
\begin{equation}
\begin{split}
	\policy^* 
	& = \argmax_{\policy \in \policySpace_{\imdp}} \, \min_{\transfunc \in \transfuncImdp} \Pr(\imdp(\policy, \transfunc) \models \propertyMDP),
	\label{eq:optimalPolicy_robust}
\end{split}
\end{equation}
where we denote by $\imdp(\policy, \transfunc)$ the Markov chain induced by \gls{iMDP} $\imdp$ under policy $\policy \in \policySpace_{\imdp}$ and by fixing $P \in \transfuncImdp$.
\begin{remark}
A reach-avoid property can alternatively be expressed using rewards, where we assign a reward of one to the goal states and zero elsewhere~\cite{DBLP:books/daglib/BaierKatoen2008}.
Here, we directly compute reach-avoid probabilities and omit rewards for brevity.
\end{remark}

%% file: 3-Filtering.tex
\section{Gaussian Belief Dynamical System}
\label{sec:BeliefSpacePlanning}
To solve \cref{prob:Problem}, we need to provide guarantees on the progression of the state $\bm{x}_{k}$ of \gls{LTI} system $\mathcal{S}$, which is not directly observable.
Instead, we use recursive state filtering techniques to update a belief over the state at each time $k$.
We make the following assumption on the initial belief $\belief(\bm{x}_0)$:
\begin{assumption}
\label{assumption:InitialBelief}
The state $\bm{x}_0$ at time $k=0$ is a Gaussian random variable distributed by $\bm{x}_0 \sim \belief(\bm{x}_0) = \Gauss[\bm\mu_0]{\cov_0}$, where $\bm\mu_0$ and $\cov_0$ are the initial mean and covariance matrix.
\end{assumption}
The Kalman filter is a widely used technique for implementing a recursive Bayes filter~\cite{Kalman1960,welch1995introduction}.
Instead of tracking the full history of measurements $\bm{y}_{1:k}$ and inputs $\bm{u}_{1:k}$ as in \cref{def:Belief}, the Kalman filter recursively updates the belief at each time $k$ based on only the current control input $\bm{u}_k$ and the obtained measurement $\bm{y}_{k+1}$.
If the prior belief $\belief(\bm{x}_k)$ of the state is Gaussian, then the posterior belief $\belief(\bm{x}_{k+1})$ is Gaussian as well~\cite{DBLP:books/daglib/Thrun2005}.
As a result, the recursive filter update computations are guaranteed to be tractable over any finite number of steps, as characterized by the following definition.
\begin{definition}[Kalman filter~\cite{DBLP:books/daglib/Thrun2005}]
\label{def:KalmanFilter}
For an \gls{LTI} system $\mathcal{S}$ with a belief $\bm{x}_k \sim \belief(\bm{x}_k) = \Gauss[\bm\mu_k]{\cov_k}$ at time $k$, the belief $\belief(\bm{x}_{k+1}) = \Gauss[\bm\mu_{k+1}]{\cov_{k+1}}$ at time $k+1$ is computed as
\begin{subequations}
    \begin{align}
    \bm\mean_{k+1} &= \hat{\bm\mean}_{k+1} + K_{k+1} (\bm{y}_{k+1} - C\hat{\bm{\mean}}_{k+1})
    \label{eq:KalmanFilter_mean}
    \\
    \cov_{k+1} &= (I_n - K_{k+1} C)(A\cov_k A^\top + \cov_{w_k}),
    \label{eq:KalmanFilter_covariance}
    \end{align}
    \label{eq:KalmanFilter}
\end{subequations}
where $\hat{\bm\mean}_{k+1}$ and $K_{k+1}$ are defined as
\small
\begin{align*}
    \hat{\bm\mean}_{k+1} &= A \bm\mean_{k} + B \bm{u}_{k} + \bm\mean_{w_k}
    \\
    K_{k+1} &= (A\cov_k A^\top + \cov_{w_k}) C^\top (C (A\cov_k A^\top + \cov_{w_k}) C^\top + \cov_{v_k})^{-1}.
\end{align*}
\end{definition}

\begin{remark}
    \label{remark:KalmanOptimality}
    For \gls{LTI} systems with additive Gaussian noise, the Kalman filter is an \emph{optimal state estimator} in the minimum mean-square-error sense, meaning its estimate is the least uncertain of any filter given the same history of information.
\end{remark}
We refer to~\cite{DBLP:books/daglib/Thrun2005, DBLP:journals/siamrev/HumpherysRW12} for a formal proof of the optimality of Kalman filters for \gls{LTI} systems.
The covariance update $\cov_{k+1}$ in \cref{eq:KalmanFilter_covariance} is a function of only the current covariance $\cov_k$ and the properties of the dynamical system $\mathcal{S}$ defined by \cref{eq:LTI_system}.
Thus, we make the following important remark:
\begin{remark}
    \label{remark:KalmanControlIndependence}
    The belief covariance $\cov_{k+1}$ is \emph{deterministic} and can, therefore, be computed a-priori from $\cov_0$ for any $k \in \N$.
\end{remark}
As the measurement $\bm{y}_{k+1}$ is only observed at time $k+1$, the belief mean $\bm\mean_{k+1}$ in \cref{eq:KalmanFilter_mean} is a \emph{random variable} at time $k$.
Therefore, the progression of the belief mean $\bm\mean_{k+1}$ can (at each step $k$) be interpreted as a stochastic dynamical system.
\begin{lemma}
    \label{lemma:BeliefDynamics}
    The mean $\bm\mean_{k+1}$ of the belief evolves according to the following stochastic dynamical system, denoted by $\mathcal{B}$:
    \begin{empheq}[left=\mathcal{B} \colon \empheqlbrace]{align}
        \, &\bm\mean_{k+1} = A \bm\mean_{k} + B \bm{u}_{k} + \bm\mean_{w_k} + K_{k+1} \bm\mean_{v_{k+1}} + \delta_{k+1},
        \hspace{-1em}
        \nonumber
        \\
        \, &\belief(\bm{x}_0) = \Gauss[\bm\mean_0]{\cov_0} \in \BeliefSpace
        \label{eq:lemma:Belief_dynamics}
    \end{empheq}
    where the \emph{belief noise} $\delta_{k+1} \sim \Gauss[0]{\cov_{\delta_{k+1}}}$ is a Gaussian random variable with zero mean and covariance
    \small
    \begin{equation}
    \cov_{\delta_{k+1}} = K_{k+1} \left( C (A \cov_{k} A^\top + \cov_{w_k}) C^\top + \cov_{v_{k+1}} \right) K_{k+1}^\top.
    \label{eq:lemma:Belief_dynamics_cov}
    \end{equation}
\end{lemma}
\begin{proof}
    By plugging in the definition of $\bm{y}_{k+1} = C \bm{x}_{k+1} + \bm{v}_{k+1}$ in \cref{eq:KalmanFilter_mean}, we obtain that
    \begin{equation}
        \bm\mean_{k+1} = \hat{\bm\mean}_{k+1} + K_{k+1} ( C(\bm{x}_{k+1} - \hat{\bm\mean}_{k+1}) + \bm{v}_{k+1} ).
        \label{eq:lemma:BeliefDynamics:proof1}
    \end{equation}
    Observe that $\bm{x}_{k+1} - \hat{\bm\mean}_{k+1}$ is a random variable distributed as
    \begin{equation}
        \bm{x}_{k+1} - \hat{\bm\mean}_{k+1} \sim \Gauss[0]{A \cov_k A^\top + \cov_{w_k}},
        \label{eq:lemma:BeliefDynamics:proof2}
    \end{equation}
    and thus, the following terms in \cref{eq:lemma:BeliefDynamics:proof1} are distributed as
    \begin{equation}
        K_{k+1} ( C(\bm{x}_{k+1} - \hat{\bm\mean}_{k+1}) + \bm{v}_{k+1} ) \sim \Gauss[ \bm\mean_{v_{k+1}} ]{ \cov_{\delta_{k+1}} },
        \label{eq:lemma:BeliefDynamics:proof3}
    \end{equation}
    where $\cov_{\delta_{k+1}}$ is defined by \cref{eq:lemma:Belief_dynamics_cov}.
    Using this result in \cref{eq:lemma:BeliefDynamics:proof1} and writing the mean $\bm\mean_{v_{k+1}}$ outside the Gaussian yields
    \begin{equation}
        \bm\mean_{k+1} = \hat{\bm\mean}_{k+1} + K_{k+1} \bm\mean_{v_{k+1}} + \delta_{k+1},
        \label{eq:lemma:BeliefDynamics:proof4}
    \end{equation}
    with $\delta_{k+1}$ as defined in \cref{lemma:BeliefDynamics}.
    Finally, by expanding \cref{eq:lemma:BeliefDynamics:proof4} with the definition of $\hat{\bm\mean}_{k+1} = A \bm\mean_{k} + B \bm{u}_{k} + \bm\mean_{w_k}$ in \cref{def:KalmanFilter}, we obtain \cref{eq:lemma:Belief_dynamics} and conclude the proof.
\end{proof}
\cref{lemma:BeliefDynamics} carries an important message: the evolution of the mean of the belief $\bm\mean_k$ (which lives on the same state space as $\bm{x}_k$) can be interpreted as a \emph{fully observable} \gls{LTI} system with additive Gaussian noise $\delta_{k+1} \sim \Gauss[0]{\cov_{\delta_{k+1}}}$.

\subsection{Augmented property}
\label{subsec:AugmentedProperty}
Like we write $\mathcal{S}(\controller)$ for the \gls{LTI} system closed under controller $\controller$, we denote by $\mathcal{B}(\controller)$ the closed-loop belief system defined by \cref{eq:lemma:Belief_dynamics}.
We will now formally relate the satisfaction probability $\Pr(\mathcal{S}(\controller) \models \pproperty)$ of the \gls{LTI} system with the satisfaction probability $\Pr(\mathcal{B}(\controller) \models \propertyAugm)$ of the belief dynamics, where $\propertyAugm$ is a modified (called \emph{augmented}) property that we define below.
The key idea is to account for the covariance $\cov_k$ between the belief mean and the actual state by \emph{expanding} critical regions and \emph{contracting} goal regions at every time $k$.
\begin{definition}
    We define by $\criticalRegion^\varepsilon$ an $\varepsilon$-expanded version of set $\criticalRegion$, and by $\goalRegion^\varepsilon$ an $\varepsilon$-contracted version of set $\goalRegion$:
    \begin{equation}
    \begin{split}
        \criticalRegion^\varepsilon &\supseteq 
        \left\{ \bm{x} \in \R^n \,\,\Big\vert\,\, \min_{\bm{x}' \in \criticalRegion} || \bm{x} - \bm{x}' ||_2 \leq \varepsilon \right\}
        \\
        \goalRegion^\varepsilon &\subseteq 
        \left\{ \bm{x} \in \R^n \,\,\Big\vert\,\, \min_{\bm{x}' \in \R^n \setminus \goalRegion} || \bm{x} - \bm{x}' ||_2 \geq \varepsilon \right\}.
    \end{split}
    \end{equation}
\end{definition}
In our experiments, we use reach-avoid properties with (unions of) rectangular goal and critical regions.
As also shown by \cref{fig:Overview}, one easy way to obtain an $\varepsilon$-expanded critical region is then to increase the halfwidth of each rectangle by $\varepsilon$ (for an $\varepsilon$-contracted goal region, we decrease the halfwidth).

We use expanded critical and contracted goal state sets to introduce the notion of an \emph{augmented} property:
\begin{definition}
    \label{def:PropertyAugmented}
    The augmented version of a reach-avoid property $\pproperty$ is a tuple $\propertyAugm = ((\goalRegion^{\varepsilon_k})_{k=0}^N, (\criticalRegion^k)_{k=0}^N, \bm{x}_0, N)$, where $(\goalRegion^{\varepsilon_k})_{k=0}^N$ and $(\criticalRegion^k)_{k=0}^N$ are time-varying sequences of contracted goal and expanded critical regions.
\end{definition}
For every $k=0,\ldots,N$ we choose $\varepsilon_k$ such that, if $\bm\mean_{k}$ is \emph{not} in the expanded critical region $\criticalRegion^{\varepsilon_k}$, then the probability for the actual state $\bm{x}_{k}$ to \emph{not} be in the (non-expanded) critical region $\criticalRegion$ is at least $\beta$.
Similarly, if $\bm\mean_{k} \in \goalRegion^{\varepsilon_k}$, then the probability for $\bm{x}_{k} \in \goalRegion$ is at least $\beta$ for every $k=0,\ldots,N$.
We find such a value for $\varepsilon_k$ by solving the following optimization program:
\begin{align}
    \minimize_{\varepsilon_k \in \R_{\geq 0}} \ & \varepsilon_k
    \label{eq:ErrorBounds}
    \nonumber
	\\
	\text{subject to} \enskip & \amsmathbb{P} \Big( \bm{z} \in 
    [-\varepsilon_k, \varepsilon_k]^n \Big) \geq \beta
	\\
	& \bm{z} \sim \Gauss[0]{\cov_k},
	\nonumber
\end{align}
where $\beta \in (0,1)$ is a confidence parameter, and $[-\varepsilon_k, \varepsilon_k]^n$ is a zero-centered hyperrectangle, with $n$ the dimension of the state.
In practice, we compute a feasible (but generally suboptimal) solution to \cref{eq:ErrorBounds} by iteratively increasing $\varepsilon_k$ until the probabilistic constraint is satisfied (which we check numerically using the method in~\cite{Cunningham2011gaussian}).
For any solution $\varepsilon_k$ to \cref{eq:ErrorBounds}, it holds that
\begin{equation}
\begin{split}
    \amsmathbb{P}(\bm{z} \notin \criticalRegion \, \vert \, \bm\mean_k \notin \criticalRegion^{\varepsilon_k}) & \geq \beta
    \\
    \amsmathbb{P}(\bm{z} \in \goalRegion \, \vert \, \bm\mean_k \in \goalRegion^{\varepsilon_k}) & \geq \beta.
\end{split}
\label{eq:AugmentedRegions}
\end{equation}
Contracting goal and expanding critical regions by the error bound $\varepsilon_k$ is also shown in \cref{fig:Overview} (middle figure).
Note that we will define the finite-state abstraction (discussed in \cref{sec:FilterBasedAbstraction}) based on these contracted and expanded regions.

As a key result, we relate the satisfaction probability of \gls{LTI} system $\mathcal{S}(\controller)$ to that of the corresponding belief system $\mathcal{B}(\controller)$:

\begin{theorem}
    \label{thm:BeliefCorrectness}
    Given a closed-loop \gls{LTI} system $\mathcal{S}(\controller)$ and a property $\pproperty$, let $\mathcal{B}(\controller)$ be the closed-loop belief system and $\propertyAugm$ the augmented property, where $\varepsilon_k$ is an optimal solution to \cref{eq:ErrorBounds} for every $k=0,\ldots,N$.
    Then, it holds that
    \begin{equation*}
        \Pr(\mathcal{S}(\controller) \models \pproperty) 
        \geq
        \Pr(\mathcal{B}(\controller) \models \propertyAugm) - (1-\beta)(N+1)
    \end{equation*}
\end{theorem}

\begin{proof}
    Observe that a sufficient condition for $\mathcal{S}(\controller) \models \pproperty$ to hold is that $\mathcal{B}(\controller) \models \propertyAugm$ and $||\bm{x}_k - \bm\mean_k ||_\infty \leq \varepsilon_k$ for all $k \in \{0,\ldots,N\}$.
    For brevity, denote by $\Gamma_k$ the event that $||\bm{x}_k - \bm\mean_k ||_\infty \leq \varepsilon_k$.
    In other words, we have that
    \begin{equation}
        \Pr(\mathcal{S}(\controller) \models \pproperty) \geq \Pr \big(
        \mathcal{B}(\controller) \models \propertyAugm \cap [\cap_{k=0}^N \Gamma_k]
        \big).
        \label{thm:BeliefCorrectness:proof1}
    \end{equation}
    For any finite collection of (possibly dependent) events $\{ \mathcal{A}_1, \ldots, \mathcal{A}_z \}$, and their complements $\{ \mathcal{A}'_1, \ldots, \mathcal{A}'_z \}$, we know via Boole's inequality that
    \begin{equation}
    \begin{split}
        \Pr(\cap_{i=1}^z \mathcal{A}_i) 
        &= 1 - \Pr( \cup_{i=1}^z \mathcal{A}'_i  )
        \geq 1 - \sum_{i=1}^z \Pr ( \mathcal{A}'_i ).
        \label{eq:UnionBound}
    \end{split}
    \end{equation}
    Thus, we rewrite \cref{thm:BeliefCorrectness:proof1} as
    \begin{align}
         \Pr(\mathcal{S}(\controller) \models \pproperty) 
         &\geq
         \Pr (\mathcal{B}(\controller) \models \propertyAugm) - \Pr(\cup_{k=0}^N \Gamma'_k)
         \nonumber
         \\
         &\geq
         \Pr (\mathcal{B}(\controller) \models \propertyAugm) - \sum_{k=0}^N \Pr(\Gamma'_k)
         \label{thm:BeliefCorrectness:proof2}
         \\
         &\geq 
         \Pr (\mathcal{B}(\controller) \models \propertyAugm) - (1-\beta)(N+1)
         \nonumber,
    \end{align}
    where we used that for every $k=0,\ldots,N$, it holds that $\Pr(\Gamma_k) \geq \beta$.
    This concludes the proof.
\end{proof}

Intuitively, \cref{thm:BeliefCorrectness} provides a lower bound on the probability that $\mathcal{S}(\controller)$ satisfies $\pproperty$, based on the probability that $\mathcal{B}(\controller)$ satisfies the augmented property $\propertyAugm$.
By choosing the hyperparameter $\beta$ sufficiently close to one, e.g., $\beta = 0.99$, we can control the tightness of this upper bound.

%% file: 4-Abstraction.tex
\section{Filter-based abstraction}
\label{sec:FilterBasedAbstraction}

In this section, we proceed with computing a bound on the satisfaction probability $\Pr(\mathcal{B}(\controller) \models \propertyAugm)$ for the belief system $\mathcal{B}$.
Specifically, our approach is to construct a finite-state abstraction of the belief system $\mathcal{B}$ as an \gls{iMDP}.
In this \gls{iMDP}, actions are associated with executing a control input $\bm{u}_k$, and the probabilistic transitions capture the belief update, which is stochastic due to the process and measurement noise.

\begin{remark}
One may intuitively think that the belief update only depends on the measurement noise.
However, a quick inspection of \cref{lemma:BeliefDynamics} reveals that the belief update depends on both the process (noise $\bm{w}_{k}$) and measurement noise ($\bm{v}_{k+1}$).
\end{remark}

\subsection{Belief discretization}
Recall that the state $\bm{x}_k$ of \gls{LTI} system $\mathcal{S}$ and the mean $\bm\mean_k$ of belief system $\mathcal{B}$ both live on the same state space, i.e., $\bm{x}_k, \bm\mean_k \in \R^n$.
We choose a \emph{partition} $\Partition$ of a compact subset $\cStateSpace \subset \R^n$ into a finite set of disjoint \emph{regions}:
\begin{definition}
    \label{def:Partition}
    A \emph{partition} $\Partition = (R_1, \ldots, R_{|\Partition|})$ of $\cStateSpace$ is a finite collection of subsets, such that the following holds:
    \begin{enumerate}
        \item $\bigcup_{i=1}^{|\Partition|} R_i = \cStateSpace$,
        \item $R_i \bigcap R_j = \emptyset, \,\, \forall i,j \in \{1,\ldots,|\Partition|\}, \,\, i \neq j$.
    \end{enumerate}
\end{definition}
The set $\cStateSpace$ is the portion of $\R^n$ that we will capture in our abstraction.
We consider the regions in $\Partition$ to be $n$-dimensional convex polytopes.
Thus, each region $R_i \in \Partition$ is the solution set of $m$ linear inequalities parameterized by $M_i \in \R^{m \times n}$ and $\bm{b}_i \in \R^m$, i.e.,
%
$
    R_i = \big\{ \bm{x} \in \R^n \, \vert \, M_i \bm{x} \leq \bm{b}_i \big\}.
$

\subsection{Interval MDP abstraction}
\label{sec:FilterBasedAbstraction_abstraction}
We formalize the dynamics of belief system $\mathcal{B}$, i.e., the evolution of the belief mean $\bm\mean_k$, as an \gls{iMDP} $\imdp = \iMDP$ by defining its states, actions, and probability intervals.

\subsubsection{States}
We define an \gls{iMDP} state $s_i^k$ for every region $R_i$ at every time step $k \in \{0,\ldots,N\}$, which represents all belief means $\bm{\mean}_k \in R_i$.
In addition, we define one absorbing state $s_\star$ and one goal state $s_g$ (we formalize the semantics for these states below).
As such, the set of \gls{iMDP} states, with $|\States| = |\Partition| (N+1) + 2$, is:
\begin{equation}
    \States = \big\{ s_i^k \ \vert \ \forall i \in \{ 1,\ldots, \vert \Partition \vert \}, \ k \in \{ 0,\ldots,\finiteHorizon \} \big\} \cup \{ s_\star, s_g, \}.
\end{equation}
We define a function $T \colon \R^n \times \{0,\ldots,N\} \to \States$ that maps belief means $\bm{\mean}_k$ of system $\mathcal{B}$ and time steps to \gls{iMDP} states:
\begin{equation}
    \label{eq:State_relation}
    T(\bm\mean, k) =
    \begin{cases}
    s_\star \enskip \text{if\,} \bm\mean \in (\R^n \setminus \cStateSpace) \cup (\cStateSpace \cap \criticalRegion^{\varepsilon_k})
    \\
    s_g \enskip \text{if\,} \bm\mean \in \cStateSpace \cap \goalRegion^{\varepsilon_k}
    \\
    s_i^k \enskip \text{otherwise, with\,}i\text{\,such that\,} \bm\mean \in R_i.
    \end{cases}
\end{equation}
Intuitively, $T(\bm\mu, k)$ maps to the absorbing state $s_\star$ if $\bm\mu$ is either outside of $\cStateSpace$ or is contained in the expanded critical state set $\criticalRegion^{\varepsilon_k}$ at time $k$.
Similarly, $T(\bm\mu, k)$ maps to the goal state $s_g$ if $\bm\mu$ is within the contracted goal state set $\criticalRegion^{\varepsilon_k}$ at time $k$.
If neither are satisfied, $T(\bm\mu, k)$ maps to the state $s_i^k$ associated with time step $k$ and index $i \in \{1,\ldots,|\Partition|\}$ for which $\bm\mu \in R_i$.
For convenience, we also denote by $R_s \in \Partition$ the region associated with a state $s \in \States \setminus \{s_\star, s_g\}$.

\subsubsection{Actions}
\label{sssec:actions}
In our abstraction, actions do not correspond to a discretization of the control space $\cControlSpace$, as is common with abstraction methods~\cite{LSAZ21}.
Instead, each action models a desired outcome for the belief mean $\bm\mean_{k+1}$ at time $k+1$.
Formally, we define $q \in \N$ \gls{iMDP} actions, such that $\Actions = \{a_1,\ldots,a_q\}$.
Every action $a \in \Actions$ is associated with a fixed point $\bm{d}_a \in \cStateSpace$ on the continuous state space, which is a \emph{target belief mean} associated with that state. 
Without loss of generality, we define one action $a$ for every region $R \in \Partition$, such that $q = |\Partition|$, and choose the target point $\bm{d}_a$ to be the center of that region.

\begin{figure}[t!]
	\centering
        \includegraphics[]{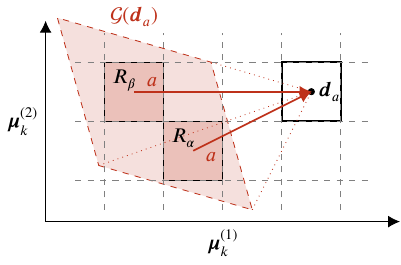}%
	\caption{
	The \gls{iMDP} action $a \in \Actions$ is only enabled in states whose region $R \in \Partition$ is contained in the backward reachable set $\mathcal{G}(\bm{d}_a)$, which (in this case) only holds for $R_\alpha, R_\beta \in \Partition$.
	}
	\label{fig:AbstractionMapping}
\end{figure}

Let us now define the semantics of \gls{iMDP} actions.
Action $a \in \Actions$ is defined such that the \emph{expected mean of the belief at time $k+1$} is equal to the target point $\bm{d}_a$ of action $a$, i.e.,
\begin{equation}
    \label{eq:Action_semantics}
    \amsmathbb{E}[\bm\mean_{k+1}] 
    = A \bm\mean_{k} + B \bm{u}_{k} + \bm\mean_{w_k} + K_{k+1} \bm\mean_{v_{k+1}} 
    = \bm{d}_a.
\end{equation}
In other words, action $a \in \Actions$ corresponds to executing a control input $\bm{u}_k$ such that $\amsmathbb{E}[\bm\mean_{k+1}] = \bm{d}_a$.
To ensure the \gls{iMDP} is a \emph{sound abstraction} of system $\mathcal{B}$, we enable action $a \in \Actions$ only in a state $s \in \States \setminus \{s_\star, s_g\}$ if, for every $\mean \in R_s$, there exists a control input $\bm{u}_k$ such that $\amsmathbb{E}[\bm\mean_{k+1}] = \bm{d}_a$.
We impose this constraint using the \emph{one-step backward reachable set} $\mathcal{G}(\bm{d}_a)$ of target mean $\bm{d}_a$~\cite{DBLP:conf/cdc/BansalCHT17}:
\begin{equation}
    \mathcal{G}(\bm{d}_a)
	= \big\{ \bm \mean \in \R^n \mid \bm{d}_a = A \bm\mean + B \bm{u}_k +      \bm{\mean}_{w_k}, 
	\,\, \bm{u}_k \in \cControlSpace \big\}.
	\label{eq:PredecessorSet}
\end{equation}
\noindent
Action $a$ exists in state $s_i^k$ if and only if $R_i \subseteq \mathcal{G}(\bm{d}_a)$.
Hence, the set $\Actions(s)$ of actions enabled in a state $s \in \States$ is
\begin{equation}
    \label{eq:EnabledActions}
    \Actions(s) = 
    \begin{cases}
        \mathrlap\varnothing\hphantom{\{ a \in \Actions \ \vert \ R_s \subseteq \mathcal{G}(\bm{d}_a) \}}\,\text{if } s \in \{s_\star, s_g\}
        \\
        \{ a \in \Actions \ \vert \ R_s \subseteq \mathcal{G}(\bm{d}_a) \}\,\,\text{otherwise.}
    \end{cases}
\end{equation}
If $A(s) = \varnothing$, we add a deterministic transition to the absorbing state $s_\star$, essentially rendering it a deadlock.
In \cref{fig:AbstractionMapping}, the set $\mathcal{G}(\bm{d}_a)$ for action $a \in \Actions$ is shown as the shaded area, so action $a$ exists in states $s_\alpha$ and $s_\beta$.

The set $\mathcal{G}(\bm{d}_a)$ can have a non-empty interior only if matrix $B$ in \cref{eq:LTI_system} has full rank, which is often not the case.
However, under the following assumption, we can always increase the rank of matrix $B$ by suitably grouping multiple discrete time steps:
\begin{assumption}
    \label{assump:PredSetRank}
    The \gls{LTI} system $\mathcal{S}$ is controllable, i.e., the controllability matrix 
    $\mathcal{C} = 
        [B \enskip AB \enskip \cdots \enskip A^{n-1}B]
    $ 
    has rank $n$.
\end{assumption}

\subsubsection{Transition probability intervals}
\label{subsec:TransitionProbabilities}
Upon choosing an action $a \in \Actions(s_i^k)$ at time $k$ in a state $s_i^k \in \States \setminus \{s_\star, s_g\}$, the expected belief mean $\amsmathbb{E}[\bm\mean_{k+1}]$ at time $k+1$ satisfies \cref{eq:Action_semantics}, and thus, the mean at time $k+1$ is written as
\begin{equation}
    \bm\mean_{k+1} \sim \Gauss[\bm{d}_a]{\cov_{\delta_{k+1}}}.
\end{equation}
Let us denote the probability density function of $\bm\mean_{k+1}$ by $p(\bm\mean_{k+1} \vert \hat{\bm{\mean}}_{k+1} = \bm{d}_a, \cov_{\delta_{k+1}})$.
The probability that action $a$ induces a transition to a belief mean $\bm\mean_{k+1}$ within some set $Z \subset \R^n$ is obtained by integrating this probability density function over that set:
\begin{align}
	F(Z, \bm{d}_a, \cov_{\delta_{k+1}})
	& =
	\int_{Z} p(\bm\mean_{k+1} \, | \, \amsmathbb{E}[\bm\mean_{k+1}] = \bm{d}_a, \cov_{\delta_{k+1}}) d\bm{\mean}_{k+1}
	\nonumber
        \\        
        &=
        \int_{Z} \Gauss[\bm{d}_a]{\cov_{\delta_{k+1}}} d\bm{\mean}_{k+1}.
        \label{eq:TransitionProb1}
\end{align}
We obtain the probabilities for a state-action pair $s_i^k \in \States \setminus \{s_\star, s_g\}$, $a \in \Actions(s_i^k)$ by replacing $Z$ with the appropriate set:
\begin{enumerate}
    \item The probability $P(s_i^k, a)(s_\star)$ to reach the absorbing state $s_\star$ is obtained for $Z \coloneqq (\R^n \setminus \cStateSpace) \cup (\cStateSpace \cap \criticalRegion^{\varepsilon_k})$;
    \item The probability $P(s_i^k, a)(s_g)$ to reach the goal state $s_g$ is obtained for $Z \coloneqq \cStateSpace \cap \goalRegion^k$;
    \item The probability $P(s_i^k, a)(s_j^{k+1})$ to reach state $s_j^{k+1}$ is obtained for $Z \coloneqq R_j \backslash (\goalRegion^k \cup \criticalRegion^k)$.
\end{enumerate}
\begin{remark}
    The sum of probabilities is $\sum_{s' \in \States} P(s_i^k, a)(s') = 1$, and is equivalent to computing $F(\R^n, \bm{d}_a, \cov_{\delta_{k+1}}) = 1$.
\end{remark}
To compute transition probabilities using \cref{eq:TransitionProb1}, we must evaluate cumulative distribution functions for multivariate Gaussians.
No closed-form expression exists for these functions, so an exact computation of these probabilities is impossible in general~\cite{Cunningham2011gaussian,Genz2000multivariateGaussians}.
Instead, we use an implementation of~\cite{Cunningham2011gaussian}, which approximates probabilities with an approximation error of below 1\%.
Thus, for every transition $(s,a,s')$, we obtain an interval $[\hat{p}-\theta, \hat{p}+\theta]$ for $\theta = 0.01$ around its point estimate $\hat{p}$ that contains the true transition probability, i.e., $P(s,a)(s') \in [\hat{p}-\theta, \hat{p}+\theta]$.
We use these intervals in the uncertain transition function $\transfuncImdp \colon \States \times \Actions \times \States \rightharpoonup \Interval$ of the \gls{iMDP}.

In summary, we construct the following abstract \gls{iMDP} of the belies system $\mathcal{B}$ defined by \cref{lemma:BeliefDynamics}:

\begin{definition}[Filter-based \gls{iMDP}]
    \label{def:FBA}
    The abstraction of the belief system $\mathcal{B}$ is an \gls{iMDP} $\imdp = \iMDP$, where
    \begin{itemize}
        \item $\States = \big\{ s_i^k \colon
        i=1,\ldots,|\Partition|,
        k=0,\ldots,\finiteHorizon
        \big\} \cup \{ s_\star, s_g \}$ is a finite set of states;
    	\item $\Actions = \big\{ a_1, a_2, \ldots, a_{|\mathcal{\Partition}|} \big\}$ is a set of actions, each with a fixed target mean $\bm{d}_a$;
    	\item $\initState = T(\bm\mean_0, 0)$ is the initial state;
    	\item $\transfuncImdp \colon \States \times \Actions \times \States \rightharpoonup \Interval \cup \{ [0,0] \}$ is the uncertain transition function, where each $\transfuncImdp(s,a,s') = [\hat{p}(s,a)(s') - \theta, \hat{p}(s,a)(s') + \theta]$ is its approximation from \cref{eq:TransitionProb1} plus-minus $\theta=0.01$.
    \end{itemize}
\end{definition}

\subsection{Controller refinement}
\label{subsec:ControlLaw}

By construction, we can refine any policy $\policy \in \policySpace_{\imdp}$ for the abstract \gls{iMDP} $\imdp$ into a controller of the form in \cref{def:controller}.
Concretely, this refined controller is obtained as follows.
\begin{definition}[Refined controller]
    \label{def:RefinedController}
    Let $\policy \in \policySpace_{\imdp}$ be any policy for the \gls{iMDP} abstraction $\imdp$. 
    The refined controller $\controller \colon \R^n \times \{0,\ldots,\finiteHorizon\} \to \cControlSpace$ for this policy is defined as
    \begin{equation}
        \controller(\bm\mean, k) = B^+ (\bm{d}_a - A\bm\mean_k - \bm\mean_{w_k} - K_{k+1} \bm\mean_{v_{k+1}}),
        \label{eq:RefinedController:law}
    \end{equation}
    where $\bm{d}_a$ is the target point associated with the action $a = \policy( T(\bm\mean,k) )$ under policy $\policy$ in \gls{iMDP} state $T(\bm\mean,k) \in \States$.
\end{definition}
The refined controller is \emph{piecewise linear} in the state $\bm{x}_k$: within each element $R_i \in \Partition$ of the partition, the target point $\bm{d}_a$ of the optimal action is constant, yielding the linear control law in \cref{eq:RefinedController:law}.
By definition of the backward reachable set in \cref{eq:PredecessorSet}, for any $\policy \in \policySpace_{\imdp}$ the refined controller $\controller$ is well-defined, i.e., for all $\bm\mean \in \R^n$ and $k \in \{0,\ldots,N\}$, we have
\begin{equation}
    B^+ (\bm{d}_a - A\bm\mean_k - \bm\mean_{w_k} - K_{k+1} \bm\mean_{v_{k+1}}) \in \cControlSpace.
\end{equation}
Moreover, observe that using controller $\controller$ in the belief system defined by \cref{lemma:BeliefDynamics}, we indeed obtain
\begin{equation}
\begin{split}
    \label{eq:Action_semantics_confirmed}
    \amsmathbb{E}[\bm\mean_{k+1}] 
    &= A \bm\mean_{k} + B \controller(\bm\mean_k, k) + \bm\mean_{w_k} + K_{k+1} \bm\mean_{v_{k+1}} 
    = \bm{d}_a,
\end{split}
\end{equation}
that is, \cref{eq:Action_semantics} is indeed satisfied.

%% file: 5-Correctness.tex
\input{Includes/Algo1}

\section{Controller Synthesis Algorithm}
\label{sec:Algorithm}

In this section, we put together the ingredients from \cref{sec:BeliefSpacePlanning,sec:FilterBasedAbstraction} to provide an algorithm for solving \cref{prob:Problem}.

\subsection{Algorithm}
Our algorithm for solving \cref{prob:Problem} is presented in \cref{alg:algorithm}.
First, in steps 1-5, we apply the methods introduced in \cref{sec:BeliefSpacePlanning} to define the belief system $\mathcal{B}$ (as per \cref{lemma:BeliefDynamics}) and compute the error bound $\varepsilon_k$ for each $k \in \{0,\ldots,N\}$ (by solving \cref{eq:ErrorBounds}).
We then use these error bounds to define the augmented property as per \cref{def:PropertyAugmented} (line 5).

Second, in steps 6-15, we apply the abstraction scheme from \cref{sec:FilterBasedAbstraction}.
Given the partition $\Partition$, we define the \gls{iMDP} states $\States$ and actions $\Actions$ (line 6), and the subsets $\Actions(s) \subseteq \Actions$ of actions enabled in each state (line 7-9).
Thereafter, we compute the probability intervals for every time step $k \in \{0,\ldots,N\}$, action $a \in \Actions$, successor state $s'$ at time $k+1$, and state $s_i^k \in \{ s_i^k \colon i=1,\ldots,|\Partition| \}$ in which $a$ is enabled (lines 10-17).
Note that these probability intervals are the same for any two states $s_i, s_j \in \States$ in which $a$ is enabled, i.e. $\transfuncImdp(s_i,a,s') = \transfuncImdp(s_j,a,s') \, \forall s' \in \States$, if $a \in \Actions(s_i)$ and $a \in \Actions(s_j)$.
We then compute an optimal policy $\policy^\star$ for the \gls{iMDP} using \cref{eq:optimalPolicy_robust} (lines 18-19).
If $\policy^\star$ satisfies the condition in line 20, we return the refined controller defined by \cref{def:RefinedController}; otherwise, we return that the reach-avoid problem was unsatisfiable.

\subsection{Correctness of the algorithm}
We show the correctness of \cref{alg:algorithm} in two steps.
First, we establish that our abstraction scheme induces a so-called \emph{probabilistic simulation relation}~\cite{hermanns2011probabilistic} from the abstraction to the belief system, which implies that, under any policy, the satisfaction probability for the \gls{iMDP} is a \emph{lower bound} on that for the belief system under the refined controller.

\begin{lemma}
    \label{lemma:abstraction_correctness}
    Given a belief system $\mathcal{B}$ and an augmented reach-avoid property $\propertyAugm$, construct the \gls{iMDP} abstraction $\imdp$ using \cref{alg:algorithm}.
    For any policy $\policy \in \policySpace_{\imdp}$ and the corresponding refined controller $\controller$ obtained from \cref{def:RefinedController}, it holds that
    \begin{equation}
        \Pr(\mathcal{B}(\controller) \models \propertyAugm)
        \geq 
        \min_{\transfunc \in \transfuncImdp} \Pr(\imdp(\pi, \transfunc) \models \propertyAugmMDP).
        \label{eq:lemma:abstraction_correctness}
    \end{equation}
\end{lemma}

\begin{proof}
    Recall from \cref{def:RefinedController} that for any belief mean $\bm\mean \in \R^n$ and for any $k \in \{0,\ldots,N\}$, the distribution of $\bm\mean_{k+1}$ is
    \begin{equation}
        \bm\mean_{k+1} \sim \Gauss[\bm{d}_a]{\cov_{\delta_{k+1}}}, \quad 
        a = \policy(T(\bm\mean, k)) \in A(s).
        \label{eq:lemma:abstraction_correctness:proof1}
    \end{equation}
    Thus, controller $\controller$ indeed induces the same probability density function for $\bm\mean_{k+1}$ as used to define the transition probabilities in \cref{eq:TransitionProb1} with function $F(\cdot)$.
    Under this induced controller, for any \gls{iMDP} state $s_i^k \in \States$ and any $k \in \{0,\ldots,N\}$, we have
    \begin{equation}
    \begin{split}
        \amsmathbb{P} \{ \bm\mean_{k+1} \in R_{s'} \mid \bm\mean_k \in R_{s_i} \}
        & =
        F(R_{s'}, \bm{d}_a, \cov_{\delta_{k+1}})
        \\
        & =
        P(s_i^k, a)(s').
        \label{eq:lemma:abstraction_correctness:proof2}
    \end{split}
    \end{equation}
    \cref{eq:lemma:abstraction_correctness:proof2} shows that 
    the map $T \colon \R^n \times \{0,\ldots,N\} \to \States$ and the refined controller $\controller$ induce a \emph{probabilistic simulation relation}~\cite{hermanns2011probabilistic} from the closed-loop abstract \gls{MDP} $\mdp(\policy)$ to the closed-loop belief system $\mathcal{B}(\controller)$.
    It has been shown by~\cite{HSA17} that measurable events have equal probability under a probabilistic simulation relation, which implies that
    \begin{equation}
        \Pr(\mathcal{B}(\controller) \models \propertyAugm)
        =
        \Pr(\mdp(\pi) \models \propertyMDP),
        \label{eq:lemma:abstraction_correctness:proof3}
    \end{equation}
    where $\mdp = \MDP$ is the \gls{MDP} under the \emph{precise} transition function $P$ defined by \cref{eq:TransitionProb1} (so not their interval estimates).
    Observe that $P \in \transfuncImdp$ (i.e., every probability is contained in its interval; cf. \cref{def:iMDP}), which by definition of \cref{eq:optimalPolicy_robust} means that
    \begin{equation}
        \Pr(\mdp(\pi) \models \propertyMDP)
        \geq
        \min_{\transfunc \in \transfuncImdp} \Pr(\imdp(\pi, \transfunc) \models \propertyMDP).
        \label{eq:lemma:abstraction_correctness:proof4}
    \end{equation}
    Combining \cref{eq:lemma:abstraction_correctness:proof3,eq:lemma:abstraction_correctness:proof4} yields the desired expression in \cref{eq:lemma:abstraction_correctness}, so we conclude the proof.
\end{proof}

We now combine \cref{lemma:abstraction_correctness} with \cref{thm:BeliefCorrectness} to show the overall correctness of our \cref{alg:algorithm} for solving \cref{prob:Problem}.

\begin{theorem}
    \label{thm:Correctness}
    For \gls{LTI} system $\mathcal{S}$ and reach-avoid property $\pproperty$, construct the \gls{iMDP} abstraction $\imdp$ using \cref{alg:algorithm} and compute the optimal policy $\policy^\star$ as per \cref{eq:optimalPolicy_robust}.
    Let $\controller$ be the refined controller under $\policy^\star$ obtained from \cref{def:RefinedController}.
    Then, it holds that
    \begin{equation}
    \begin{split}
        \label{eq:thm:Correctness}
        \Pr(\mathcal{S}(\controller) \models \pproperty) 
        \geq p^\star - (1-\beta)(N+1),
    \end{split}
    \end{equation}
    where $p^\star \in [0,1]$ is the satisfaction probability for policy $\policy^\star$:
    \begin{equation}
        p^\star = \min_{\transfunc \in \transfuncImdp} \Pr(\imdp(\policy^\star, \transfunc) \models \propertyAugmMDP).
    \end{equation}
\end{theorem}

\begin{proof}
    Combining \cref{thm:BeliefCorrectness} with \cref{lemma:abstraction_correctness} yields
    \begin{align}
        \nonumber
        \Pr(\mathcal{S}(\controller) \models \pproperty) 
        \geq &
        \Pr(\mathcal{B}(\controller) \models \propertyAugm) 
        - (1-\beta)(N+1)
        \\
        \geq &
        \min_{\transfunc \in \transfuncImdp} \Pr(\imdp(\policy^\star, \transfunc) \models \propertyAugmMDP) 
        \\
        \nonumber
        & 
        - (1-\beta)(N+1),
    \end{align}
    which directly leads to the desired expression in \cref{eq:thm:Correctness}.
\end{proof}

Observe that if $p^\star - (1-\beta)(N+1) \geq \eta$, \cref{thm:Correctness} provides a solution to \cref{prob:Problem}.
If, on the other hand, the value of $p^\star$ is too low to satisfy the desired threshold $\eta \in [0,1]$, then the algorithm returns that the problem could not be solved.
In such a case, one may strengthen (increase) the confidence level $\beta$ (used to compute the error bounds $\varepsilon_k$) or try a more fine-grained partition of the state space.
Thus, our algorithm is \emph{sound} but not \emph{complete}: any controller returned by \cref{alg:algorithm} solves \cref{prob:Problem}, but failure to return a controller does not disprove the existence of such a controller.

%% file: Includes/Algo1.tex
\begin{algorithm}[t!]
\caption{Controller synthesis via filter-based abstraction.}
\label{alg:algorithm}
\textbf{Input}: \gls{LTI} system $\mathcal{S}$; reach-avoid property $\pproperty$; threshold $\eta$\\
\textbf{Params}: Partition $\Partition$; interval precision $\theta$\\
\textbf{Output}: Feedback controller $\controller$
\begin{algorithmic}[1] 
\STATE Define belief system $\mathcal{B}$ over horizon $k \in \{0, \ldots, N\}$
\FORALL{time steps $k \in \{0, \ldots, N\}$}
    \STATE Compute error bound $\varepsilon_k$ by solving \cref{eq:ErrorBounds}
\ENDFOR
\STATE Define augmented reach-avoid property $\propertyAugm$
\STATE Given $\Partition$, define \gls{iMDP} states $\States$ and actions $\Actions$
\FORALL{\gls{iMDP} states $s \in \States$}
    \STATE Compute enabled actions $\Actions(s) \subseteq \Actions$ via \cref{eq:EnabledActions}
\ENDFOR
\FORALL{time steps $k \in \{0,\ldots,N\}$}
    \FORALL{\gls{iMDP} actions $a \in \Actions$}
        \FORALL{$s' \in \{s_i^{k+1} \colon i = 1,\ldots,|\Partition| \} \cup \{ s_\star, s_g \}$}
            \STATE Compute $\hat{p}(\cdot,a)(s')$ using \cref{eq:TransitionProb1}
            \STATE $\transfuncImdp(s_i^k, a, s') = \hat{p}(\cdot,a)(s') \pm \theta \ \forall s_i^k$ s.t. $a \in \Actions(s_i^k)$
        \ENDFOR
    \ENDFOR
\ENDFOR
\STATE Generate \gls{iMDP} $\imdp=\iMDP$
\STATE Compute $\policy^*$ and $p^\star = \min_{\transfunc \in \transfuncImdp} \Pr(\imdp(\policy^*) \models \propertyAugmMDP)$
\IF{$p^\star - (1-\beta)(N+1) \geq \eta$}
    \STATE \textbf{Return} Refined controller $\controller$ based on \cref{def:RefinedController}
\ELSE
    \STATE \textbf{Return} $\mathsf{Unsatisfiable}$
\ENDIF
\end{algorithmic}
\end{algorithm}

%% file: 6-Extensions.tex
\section{Two-phase time horizon}
\label{sec:2phasehorizon}

The \gls{iMDP} defined in \cref{sec:FilterBasedAbstraction} has $\vert \States \vert = (\finiteHorizon+1) \vert \Partition \vert + 2$ states, i.e., one for every region of partition $\Partition$ at every time step, plus two for the absorbing and critical states.
Modeling time explicitly in the \gls{iMDP}'s states is necessary because the transition probabilities defined in \cref{eq:TransitionProb1} are \emph{time-varying} due to the dependence on the covariance matrix $\cov_{\delta_{k+1}}$.

\begin{figure}[t!]
	\centering
        \includegraphics[width=.9\linewidth]{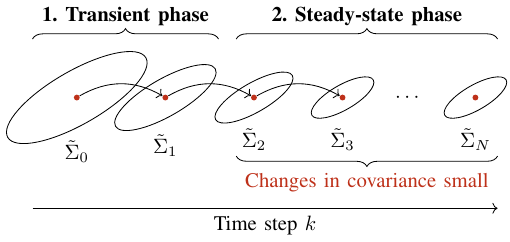}%
	\caption{Using the two-phase time horizon (shown for $\bar{\finiteHorizon} = 2$), we model the first two time steps (the transient phase) explicitly, while we model all other steps (the transient phase) together by using more conservative probability intervals.}
	\label{fig:2PhaseHorizon}
\end{figure}

However, as the distributions of the process noise $\bm{w}_k$ and measurement noise $\bm{v}_k$ are constant, the covariance matrix $\cov_{\delta_{k+1}}$ will converge in the limit~\cite{welch1995introduction}.
In practice, this convergence happens in just a few time steps (\eg, 3 or 4, as observed in our experiments in \cref{sec:NumericalStudy}).
To take advantage of this converging behavior and reduce the size of the abstract \gls{iMDP}, we propose to divide the time horizon of $\finiteHorizon$ steps into two \emph{phases}, as shown in \cref{fig:2PhaseHorizon}.
First, in the \emph{transient phase}, which ranges between steps $ k \in \{ 0,\ldots,\bar\finiteHorizon-1 \}$, with $\bar{\finiteHorizon} < N$, we model every step $k$ explicitly as before.
Thereafter, the \emph{steady-state phase} of steps $k \in \{\bar\finiteHorizon, \ldots, \finiteHorizon \}$, is modeled as a single step in the \gls{iMDP}.
Thus, the number of \gls{iMDP} states is reduced to $|\States| = (\bar\finiteHorizon+1)\vert\Partition\vert + 2$, where $\bar{\finiteHorizon} < \finiteHorizon$.

\subsection{Modeling the steady-state phase}
To implement the two-phase time horizon, we alter the set of \gls{iMDP} (as defined in \cref{sec:FilterBasedAbstraction_abstraction}) as follows:
\begin{equation}
    \label{eq:steady-state:states}
    \overline\States = \big\{ s_i^k \ \vert \ \forall i \in \{ 1,\ldots, \vert \Partition \vert \}, \ k \in \{ 0, \ldots, \bar{\finiteHorizon} \} \big\} \cup \{ s_a, s_g, \}.
\end{equation}
That is, we define $|\Partition|$ \gls{iMDP} states for every time step in the transient phase of $k \in \{0,\ldots,\bar{N}-1\}$, and another $|\Partition|$ states for the steady-state phase $k = \bar{N}$.
Observe that the (enabled) actions remain unaffected by the two-phase time horizon.

For the transient phase, we follow the exact same procedure as in \cref{alg:algorithm} to define the error bound $\varepsilon_k$ and the transition function $\transfuncImdp$.
However, for the steady-state phase, we augment the reach-avoid property by the \emph{maximal} value for the error bound $\varepsilon_k$ over all time steps $k = \bar{\finiteHorizon}, \ldots, \finiteHorizon$, which is computed as follows:
\begin{equation}
    \varepsilon_{\bar{\finiteHorizon}} = 
    \max( \{ \varepsilon_k \,\colon\, k = \bar{\finiteHorizon}, \ldots, \finiteHorizon \} ).
    \label{eq:steady-state:epsilon}
\end{equation}
Similarly, we need to compute upper and lower bounds on the probability intervals for all time steps $k = \bar{\finiteHorizon}, \ldots, \finiteHorizon$.
Thus, we define the filter-based \gls{iMDP} with 2-phase horizon as follows.

\begin{definition}[Filter-based \gls{iMDP} with 2-phase horizon]
    \label{def:FBA_2phase}
    The abstraction of the belief system $\mathcal{B}$ with a transient phase of length $\bar{\finiteHorizon} < \finiteHorizon$ is an \gls{iMDP} $\imdp^{\bar{\finiteHorizon}} = (\overline{\States}, \Actions, \initState, \overline{\transfuncImdp})$ where $\overline{\States}$ is defined by \cref{eq:steady-state:states}, $\Actions$ and $\initState$ are defined as in \cref{def:FBA}, and the transition function $\overline\transfuncImdp \colon \States \times \Actions \times \States \rightharpoonup \Interval \cup \{ [0,0] \}$ is:
    \begin{equation}
        \overline{\transfuncImdp}(s,a,s') = \begin{cases}
            \transfuncImdp(s,a,s') & \text{ if } s \in \{s_i^k \colon k < \bar{\finiteHorizon}\}
            \\
            \transfuncImdp^+(s,a,s') & \text{ otherwise},
        \end{cases}
    \end{equation}
    where $\transfuncImdp^+(s,a,s')$ is defined as
    \begin{equation}
    \begin{split}
        \transfuncImdp^+(s_i,a,s_j)
        =
        \Big[
        &\min( \cup_{k=\bar{\finiteHorizon}}^\finiteHorizon \transfuncImdp( s_i^k, a, s_j^{k+1} ) ),
        \\
        &\quad \max( \cup_{k=\bar{\finiteHorizon}}^\finiteHorizon \transfuncImdp( s_i^k, a, s_j^{k+1} ) )
        \Big].
    \end{split}
    \end{equation}
\end{definition}
In other words, the probability interval for each transition $(s,a,s')$ in the steady-state phase $k=\bar{\finiteHorizon}$, is computed as the \emph{smallest interval that contains all intervals} for that same transition at steps $k=\bar{\finiteHorizon}, \ldots, \finiteHorizon$.
We now show that the two-phase time horizon preserves the correctness of our method.

\begin{theorem}
    \label{thm:correctness:2phase}
    Let $\imdp$ and $\imdp^{\bar{\finiteHorizon}}$ be iMDP abstractions obtained for the same LTI system $\mathcal{S}$ and reach-avoid property $\pproperty$, but where the latter uses the two-phase time horizon with transient phase of length $\bar{\finiteHorizon}$.
    For any policy $\policy \in \policySpace_{\imdp}$, it holds that
    \begin{equation}
        \label{eq:correctness:2phase}
        \min_{\transfunc \in \overline{\transfuncImdp}} \Pr(\imdp^{\bar{\finiteHorizon}}(\policy, \transfunc) \models \propertyAugmMDP)
        \leq
        \min_{\transfunc \in \transfuncImdp} \Pr(\imdp(\policy, \transfunc) \models \propertyAugmMDP).
    \end{equation}
\end{theorem}

\begin{proof}
    From \cref{def:FBA_2phase}, it is straightforward to see that for every $i,j \in \{1,\ldots,|\Partition\}$ and $k \in \{\bar\finiteHorizon,\ldots,\finiteHorizon\}$, it holds that
    \begin{equation}
    \begin{split}
        \transfuncImdp(s_i^k, & \ a, s_j^{k+1}) \subseteq \overline{\transfuncImdp}(s_i^{\bar\finiteHorizon}, a, s_j^{\bar\finiteHorizon}) 
        \\
        & \enskip \forall i,j \in \{1,\ldots,|\Partition\}, \enskip
        k \in \{\bar\finiteHorizon,\ldots,\finiteHorizon\}.
    \end{split}
    \end{equation}
    That is, the probability intervals of \gls{iMDP} $\imdp$ are contained in those of $\imdp^{\bar\finiteHorizon}$.
    Thus, the lower bound on the satisfaction probability for \gls{iMDP} $\imdp^{\bar{\finiteHorizon}}$ cannot exceed that for \gls{iMDP} $\imdp$, so the claim in \cref{eq:correctness:2phase} follows.
\end{proof}

For small values of $\bar{\finiteHorizon}$, the bound in \cref{eq:correctness:2phase} will generally be loose.
Thus, the value of $\bar{\finiteHorizon}$ provides a trade-off between the \emph{size of the \gls{iMDP}}, versus the \emph{level of conservatism} of the guaranteed bound for satisfying the reach-avoid property.

%% file: 7-NumericalStudy.tex
\input{Figures/table}

\section{Numerical Experiments}
\label{sec:NumericalStudy}

Using \cref{thm:Correctness,thm:correctness:2phase}, we can compute a controller $\phi$ with a \emph{guaranteed lower bound} $p^\star - (1-\beta)(N+1)$ on the probability $\Pr(\mathcal{S}(\controller) \models \pproperty)$ that the closed-loop system satisfies property $\pproperty$.
We perform numerical experiments to answer the following questions about our approach:
\begin{enumerate}
    \item[Q1)] \change{Can our method solve \cref{prob:Problem}, and how does our method compare to sample-based planning methods?}
    \item[Q2)] How does the state space partition affect the size of abstract \glspl{iMDP} versus the guaranteed lower bounds?
    \item[Q3)] Are the guaranteed lower bound probabilities for satisfying the property indeed achieved in simulations?
    \item[Q4)] How does the two-phase time horizon control the size of abstract iMDPs vs. the quality of obtained controllers?
\end{enumerate}
To answer Q1, we consider \gls{UAV} reach-avoid problems in 2D and 3D (yielding \gls{LTI} systems of dimension $n=4$ and $6$).
To answer Q2 and Q3, we consider a partially observable variant of the package delivery benchmark from~\cite{ARCH22:ARCH_COMP22_Category_Report_Stochastic}.
Finally, to answer Q4, we consider a partially observable extension of the spacecraft rendezvous problem from~\cite{DBLP:conf/hybrid/VinodGO19}.
To compute reach-avoid probabilities and policies for \glspl{iMDP} via \cref{eq:optimalPolicy_robust}, we use an implementation of the algorithm by~\cite{DBLP:conf/cdc/WolffTM12} in the model checker \prism~\cite{DBLP:conf/cav/KwiatkowskaNP11}.
Our implementation is available at \url{\gitrepo}.
The experiments run single-threaded on a computer with a 4GHz Intel Core i9 CPU and 32 GB of RAM.
In all experiments, we compute the error bounds $\varepsilon_k$ (to expand/contract regions) using \cref{eq:ErrorBounds} for a confidence level of $\beta = 0.999$.

\subsection{Benchmark statistics}
\label{subsec:BenchmarkStatistics}
An overview of all benchmark instances is shown in \cref{tab:Benchmarks}.
The \emph{guaranteed bound} $\eta^\star_{\initState}$ is the highest lower bound on the satisfaction probability $\Pr(\mathcal{S}(\controller) \models \pproperty)$ under the refined controller (\cref{def:RefinedController}) that \cref{thm:Correctness,thm:correctness:2phase} guarantee:
\begin{equation}
    \label{eq:etaBound}
    \eta^\star_{\initState} = p^\star_{\initState} - (1-\beta)(N+1) \leq \Pr(\mathcal{S}(\controller) \models \pproperty),
\end{equation}
with $\beta = 0.999$ the confidence level, $N \in \N$ the horizon of the property, and $p^\star_{\initState} = \min_{\transfunc \in \transfuncImdp} \Pr(\imdp(\policy^\star, \transfunc) \models \propertyAugmMDP)$ the satisfaction probability under the optimal policy $\policy^\star$ computed by \cref{eq:optimalPolicy_robust} (from the initial iMDP state $\initState$ corresponding with $\bm{x}_0$).
\cref{tab:Benchmarks} shows that the bound $\eta^\star_{\initState}$ generally increases with the partition resolution and decreases with the noise strength.

We validate the correctness of the bounds $\eta^\star_{\initState}$ empirically by performing $M = 1000$ Monte Carlo simulations under the refined feedback controller.
We compute the empirical fraction $\bar{p}_{\initState} = \frac{1}{M} \sum_{i=1}^M [\omega_i \models \pproperty]$ of the trajectories satisfying the reach-avoid property, where $\omega_i = (\bm{x}_0,\bm{x}_1,\ldots,\bm{x}_N)_i$ denotes state trajectory $i \in \{1,\ldots,M\}$, and $\omega_i \models \pproperty$ is $1$ if trajectory $\omega_i$ satisfies $\pproperty$ and $0$ otherwise.
In the limit, $\bar{p}_{\initState}$ approaches the satisfaction probability $\Pr(\mathcal{S}(\controller) \models \pproperty)$ on the concrete \gls{LTI} system.
From \cref{tab:Benchmarks}, we observe that $\eta^\star_{\initState} \leq \bar{p}_{\initState}$ for all instances, \ie, the guaranteed satisfaction probability is indeed a lower bound on the empirical satisfaction probability.
This result empirically confirms the soundness of \cref{thm:Correctness,thm:correctness:2phase}.


\subsection{\gls{UAV} reach-avoid control}
\label{subsec:experiments:UAV}
Consider a \gls{UAV} reach-avoid problem in two spatial dimensions, where only the position is observed.
The dynamics are%
\begin{subequations}%
	\begin{align}%
		\bm{x}_{k+1} &= \begin{bmatrix}
			1 & 0.95 & 0 & 0 \\
			0 & 0.90 & 0 & 0 \\
			0 & 0 & 1 & 0.93 \\
			0 & 0 & 0 & 0.96
		\end{bmatrix} \bm{x}_k + \begin{bmatrix}
			0.48 & 0   \\
			0.94   & 0   \\
			0   & 0.43 \\
			0   & 0.92
		\end{bmatrix} \bm{u}_k + \bm{w}_k
		\label{eq:XY_robot_process}
		\\
		\bm{y}_{k+1} &= \begin{bmatrix} 1 & 0 & 0 & 0 \\ 0 & 0 & 1 & 0 \end{bmatrix} \bm{x}_{k+1} + \bm{v}_{k+1},
		\label{eq:XY_robot_measurement}
	\end{align}%
	\label{eq:XY_robot}%
\end{subequations}%
with process noise $\bm{w}_k \sim \Gauss[0]{f \cdot \diag{0.1, 0.02, 0.1, 0.02}}$ and measurement noise $\bm{v}_k \sim \Gauss[0]{f \cdot \diag{0.1, 0.1}}$, where $f > 0$ is the noise level.
The control input space is $\bm{u}_k \in \cControlSpace = [-4, 4]^2$.
We consider a horizon of $N=24$ steps but lump together every two time steps to satisfy \cref{assump:PredSetRank}.
The initial belief is $\bm{\mean}_0 = [-8, 0, -8, 0]$, $\cov_0 = \diag{2, 0.01, 2, 0.01}$.
We use a partition into $3\,025$ regions and the two-phase time horizon with a transient phase of $\bar{N} = 4$ steps.

\begin{figure}[t!]
    \centering
    \resizebox{\linewidth}{!}{%
    \subfloat[High noise $(f=1)$.]{%
    {\includegraphics[height=3.2cm]{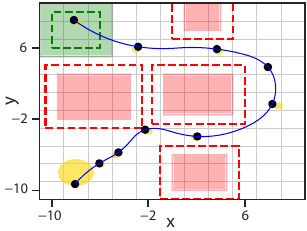}}
    \label{fig:Results_motion_highNoise} }
    \subfloat[Low noise $(f=0.1)$.]{%
    {\includegraphics[height=3.2cm]{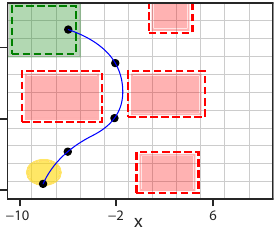}}
    \label{fig:Results_motion_lowNoise} }
    }
    \caption{
    Simulations for the 2D \gls{UAV} benchmark (showing position variables only).
    Dashed green (red) lines are the contracted goal (expanded critical) regions for the steady-state phase, and yellow ellipses show the belief covariance $\cov_k$.
    }
    \label{fig:Results_2Dmotion}
\end{figure}

\begin{figure}[t!]
    \centering
    \resizebox{\linewidth}{!}{%
    \subfloat[No input constraints.]{%
    {\includegraphics[height=3.2cm]{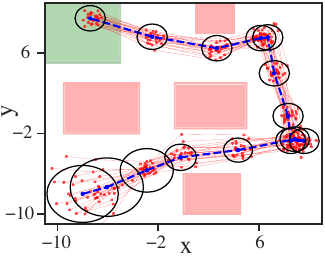}}
    \label{fig:Results_motion_RRBT_unconstrained} }
    \subfloat[With inputs constraints.]{%
    {\includegraphics[height=3.2cm]{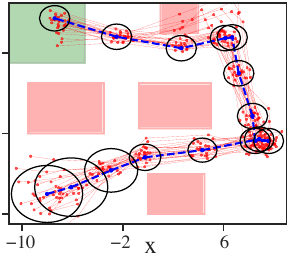}}
    \label{fig:Results_motion_RRBT_constrained} }
    }
    \caption{
    Nominal trajectories (blue dashed lines) and uncertainty predictions (black ellipses) for the RRBT on the 2D \gls{UAV} with high noise $(f=1)$.
    The red lines are simulated state trajectories, showing that the RRBT uses incorrect state uncertainty predictions in the presence of input constraints.
    }
    \label{fig:Results_2Dmotion:RRBT}
\end{figure}

\subsubsection*{Q1) Solving \cref{prob:Problem} for different noise levels}
\change{
We compare scenarios with noise levels of $f = 0.1$ and $1$.
The resulting bounds $\eta^\star_{\initState}$ on the satisfaction probability are shown in \cref{tab:Benchmarks}.
For the low-noise instances, our method provides a tight lower bound $\eta^\star_{\initState} = 0.929$ on the empirical satisfaction probability $\bar{p}_{\initState} = 1.000$.
For the high-noise instance, the lower bound ($\eta^\star_{\initState} = 0.723$) is more conservative.
\cref{fig:Results_2Dmotion} shows simulations under the refined feedback controller for both noise levels.
As expected, the error bound $\varepsilon_k$ by which critical regions are expanded (and goal regions contract) increases with the noise strength.
Under high noise, the controller chooses a longer path to navigate around the obstacles, whereas under low noise, the narrow but much shorter path to the goal is chosen.
}

\subsubsection*{Comparison to RRBT}
\change{
We now compare our method against the Rapidly-exploring Random Belief Tree (RRBT)~\cite{DBLP:conf/icra/BryR11}, a state-of-the-art sample-based method for motion planning under uncertainty.
The RRBT incrementally builds a tree of motion plans in belief space, consisting of nominal trajectories stabilized with a linear estimator and controller.
We compute the stabilizing controller using the LQR and require the collision probability to be below $\delta = 0.01$ at each step.\footnote{The code to run this experiment is in our implementation referred to earlier.}
\cref{fig:Results_2Dmotion:RRBT} shows the resulting best nominal trajectories and uncertainty predictions after $1\,000$ iterations, as well as $25$ simulated state trajectories.
Without constraints on the control input $\bm{u}_k \in \cControlSpace$, RRBT successfully finds a safe motion plan.
However, when we bound the inputs to $\cControlSpace = [-4, 4]^2$, the RRBT yields unsafe behavior (\eg, the probability of a collision at time $k=13$ is $0.22$, which is much higher than the threshold of $\delta = 0.01$).
The uncertainty predictions by the RRBT rely on a stabilizing controller that is unbounded, meaning that the resulting plan may not be feasible on the concrete \gls{LTI} system.
By contrast, the feedback controllers computed with our approach are feasible on the concrete \gls{LTI} system by construction.
}

\begin{figure}[t!] 
    \centering
    \includegraphics[width=1.05\linewidth]{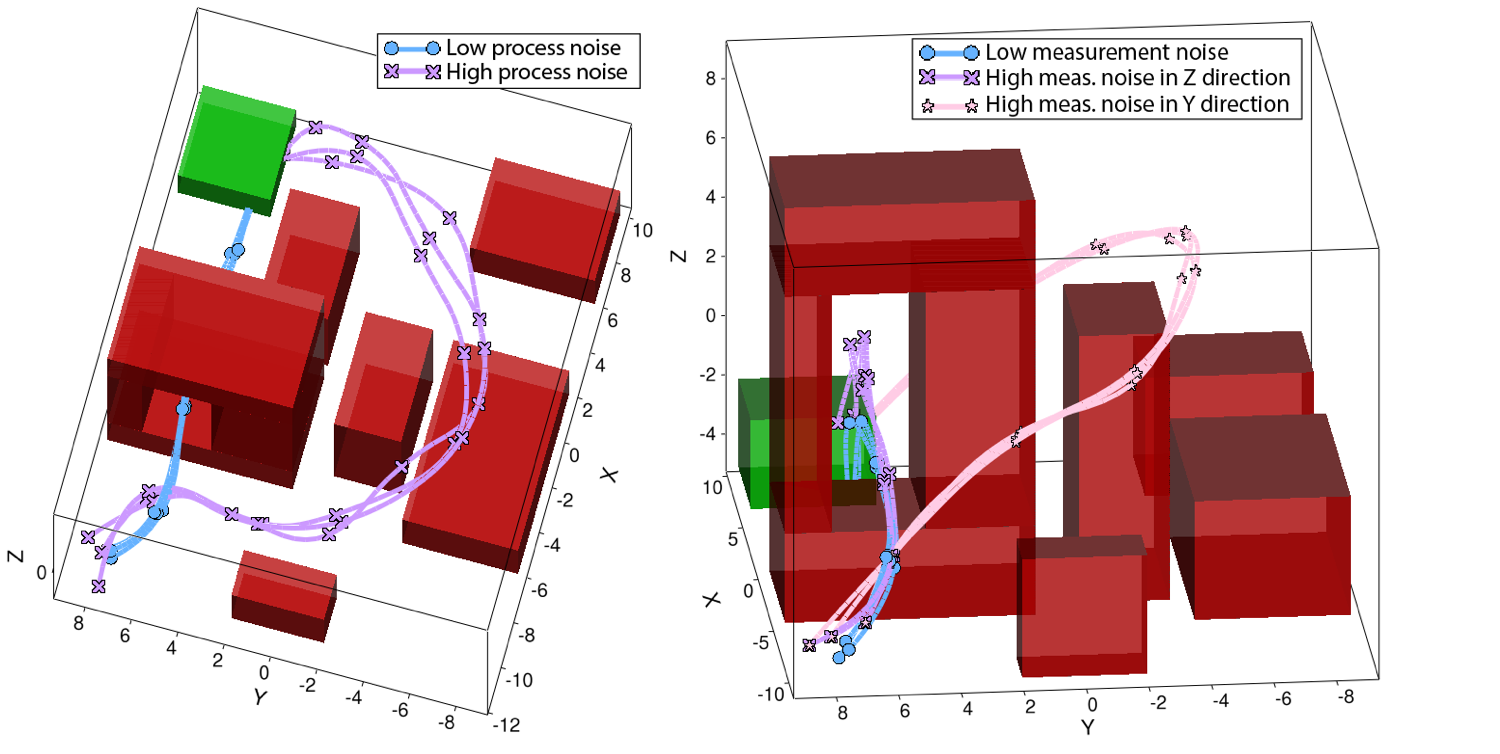}
    \caption{Simulated trajectories for the 3D \gls{UAV} benchmark with fixed measurement noise (left) and fixed process noise (right).}
    \label{fig:UAV_3D}
\end{figure}


\subsubsection*{3D \gls{UAV} benchmark}
\change{
We extend the 3D \gls{UAV} model from~\cite{Badings2022JAIR} with partial observability (referring to \cite{Badings2022JAIR} for the explicit model dynamics for brevity).
The reach-avoid problem is shown in \cref{fig:UAV_3D}, and the initial belief is $\bm{\mean}_0 = [-9.5, 0, 7.5, 0, -4, 0]$, $\cov_0 = \diag{2, 0.01, 2, 0.01, 2, 0.01}$.
We use a partition into $13\,365$ regions.
As for the 2D benchmark, our method provides tight bounds on the satisfaction probability (see \cref{tab:Benchmarks}), but these bounds become more conservative if the noise is high.
\cref{fig:UAV_3D} shows state trajectories for the 3D UAV with different noise levels.
Depending on the noise level, the \gls{UAV} either flies through the narrow pass or takes the longer path around the obstacles.
Moreover, the direction of the noise also affects the optimal path, as shown by the high measurement noise in either the $Y$ or $Z$ direction.
Thus, our method is able to synthesize correct-by-construction controllers under varying noise conditions.
}

\subsection{Package delivery}
\label{subsec:experiments:packagedelivery}
The package delivery benchmark originates from~\cite{ARCH22:ARCH_COMP22_Category_Report_Stochastic}, which we extend with measurement noise.
The model has a 2D state $\bm{x}_k \in \R^2$, control $\bm{u}_k \in [-1, 1]^2 \subset \R^2$, and dynamics%
\begin{subequations}%
	\begin{align}%
		\bm{x}_{k+1} &= \begin{bmatrix}
        0.9 & 0 \\
        0   & 0.8
    \end{bmatrix} \bm{x}_k + \begin{bmatrix}
        1.4 & 0 \\
        0   & 1.4
    \end{bmatrix} \bm{u}_k + \bm{w}_k
		\label{eq:package_process}
		\\
		\bm{y}_{k+1} &= \begin{bmatrix} 
            1 & 0 \\ 
            0 & 1 
        \end{bmatrix} \bm{x}_{k+1} + \bm{v}_{k+1},
		\label{eq:package_measurement}
	\end{align}%
	\label{eq:package}%
\end{subequations}%
with process noise $\bm{w}_k \sim \Gauss[0]{\diag{0.1, 0.1}}$ and measurement noise $\bm{v}_k \sim \Gauss[0]{\diag{0.1, 0.1}}$.
The goal is to reach the goal states $\goalRegion = [-4, -2] \times [-4, -2]$ within $N = 24$ steps, while avoiding states in $\criticalRegion = [0, 1] \times [-5, 1]$.
The initial belief is $\bm{\mean}_0 = [4.25, -4.25]$, $\cov_0 = \diag{0.5, 0.5}$.
We use \cref{alg:algorithm} to compute a feedback controller on a bounded portion $\cStateSpace = [-6, 6]^2 \subset \R^n$ of the state space, using the two-phase time horizon with a transient phase of $\bar{N} = 4$ steps.

\begin{figure}[t!]
    \centering
    \resizebox{\linewidth}{!}{%
    \subfloat[$16 \times 16$ partition.]{%
    {\includegraphics[height=3.7cm]{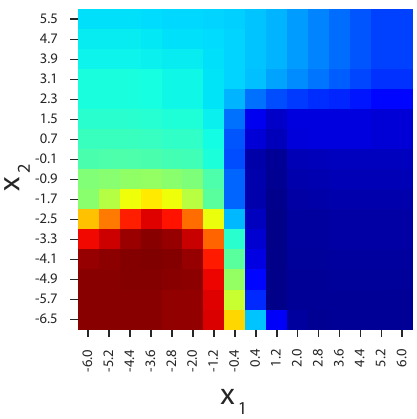}}
    \label{fig:Results:package:heatmap:16x16} }
    \subfloat[$48 \times 48$ partition.]{%
    {\includegraphics[height=3.7cm]{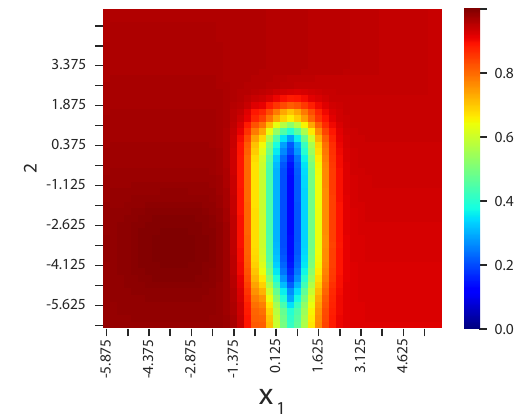}}
    \label{fig:Results:package:heatmap:48x48} }
    }
    \caption{
    Lower bounds $\eta^\star_{\initState}$ on the satisfaction probability for any initial state $\initState$, for the package delivery benchmark.
    }
    \label{fig:Results:package:heatmap}

    \vspace{1em}

    \centering
    \resizebox{\linewidth}{!}{%
    \subfloat[$16 \times 16$ partition.]{%
    {\includegraphics[height=3.7cm]{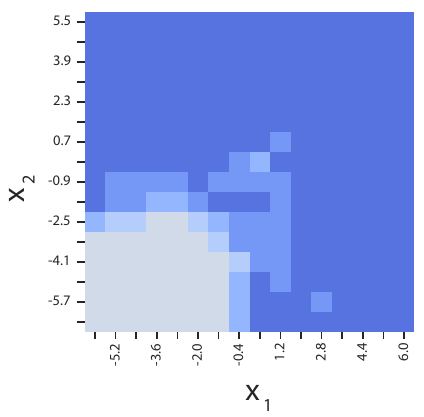}}
    \label{fig:Results:package:validation:24x24} }
    \subfloat[$48 \times 48$ partition.]{%
    {\includegraphics[height=3.7cm]{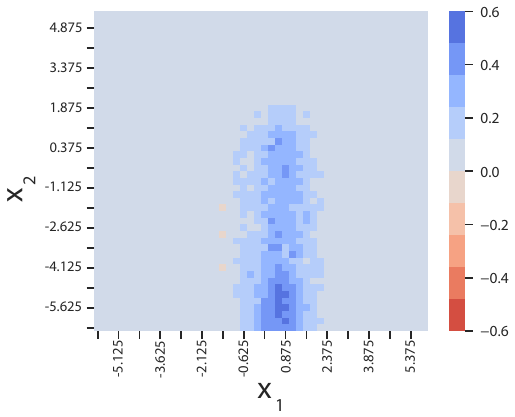}}
    \label{fig:Results:package:validation:48x48} }
    }
    \caption{
    Differences $\bar{p}_{\initState} - \eta^\star_{\initState}$ between the empirical satisfaction probability $\bar{p}_{\initState}$ and the guaranteed satisfaction probabilities $\eta^\star_{\initState}$ for every initial state $\initState$, for package delivery benchmark.
    }
    \label{fig:Results:package:validation}
\end{figure}

\subsubsection*{Q2) Partition resolution}
As shown in \cref{tab:Benchmarks}, the resolution of the state space partition provides a trade-off between the abstraction size (and thus the computational complexity) and the control precision.
In \cref{fig:Results:package:heatmap}, we show heatmaps of the lower bound satisfaction probabilities $\eta^\star_{\initState}$ from every initial \gls{iMDP} state $\initState = s_i^0$, with $i \in \{1,\ldots,|\Partition|\}$, for two partition resolutions.
For this benchmark, the partition into $16 \times 16$ regions is too coarse to obtain a representative abstraction, leading to a controller with poor (low) satisfaction guarantees.
On the other hand, the $48 \times 48$ partition does yield a feedback controller with strong satisfaction guarantees (except when starting in an initial state coinciding with the critical region).

\subsubsection*{Q3) Tightness of bound $\eta^\star_{\initState}$}
To further investigate the tightness of the lower bounds $\eta^\star_{\initState}$, we repeat the Monte Carlo simulations (as described in \cref{subsec:BenchmarkStatistics}) for every initial state.
\cref{fig:Results:package:validation} shows the values of $\bar{p}_{\initState} - \eta^\star_{\initState}$ for every initial state, i.e., the empirical satisfaction probability minus the lower bounds guaranteed by \cref{thm:Correctness,thm:correctness:2phase}.
\cref{fig:Results:package:validation} shows that we have $p^\star_{\initState} \geq \eta^\star_{\initState}$ for all $\initState$, thus confirming that our method is sound.
While the satisfaction \emph{guarantees} for the $16 \times 16$ partition are poor, the \emph{empirical satisfaction probability} $\bar{p}_{\initState}$ of the controller is still reasonably good.
For the $48 \times 48$ partition, our algorithm returns more conservative bounds on the satisfaction probabilities near the boundaries of obstacles due to the expansion of the critical regions.
In other regions of the state space, the bounds are reasonably tight.

\subsection{Spacecraft rendezvous problem}
\label{subsec:experiments:spacecraft}
We consider a variant of the spacecraft rendezvous problem supplied with the MATLAB toolbox SReachTools~\cite{DBLP:conf/hybrid/VinodGO19}, an optimization-based toolbox for probabilistic reachability problems.
The problem is to navigate one spacecraft to another while avoiding a set of obstacles.
The 4D state $\bm{x} = [p_x, p_x, v_x, v_y]^\top \in \R^4$ describes the position and velocity in both directions.
We extend the discrete-time dynamics used in~\cite{DBLP:conf/hybrid/VinodGO19} with partial observability as follows:
\begin{subequations}
	\begin{align}
		\bm{x}_{k+1} =& \begin{bmatrix}
			1.0006 & 0.0000 & 19.9986 & 0.4100 \\
			\num{8.62e-6} & 1.0000 & -0.4100 & 19.9944 \\
			\num{6.30e-5} & 0.0000 & 0.9998 & 0.0410 \\
			\num{-1.29e-6} & 0.0000 & -0.0410 & 0.9992
		\end{bmatrix} \bm{x}_k
        \nonumber
        \\ &
        + \begin{bmatrix}
			0.6666   & 0.0091   \\
			-0.0091  & 0.6666   \\
			0.0666   & 0.0014 \\
			-0.0014  & 0.0666
		\end{bmatrix} \bm{u}_k + \bm{w}_k
		\label{eq:XY_shuttle_process}
        \\
    \bm{y}_{k+1} =& \begin{bmatrix} 1 & 0 & 0 & 0 \\ 0 & 1 & 0 & 0 \end{bmatrix} \bm{x}_{k+1} + \bm{v}_{k+1},
		\label{eq:XY_shuttle_measurement}
	\end{align}%
	\label{eq:XY_shuttle}%
\end{subequations}%
with control input space $\cControlSpace = [-2.5, 2.5]^2$.
We remark that SReachTools is limited to fully observable systems and convex safe sets, which makes a direct comparison not possible.

We consider a reach-avoid problem with the same layout as in \cref{fig:Results_2Dmotion} with a horizon of $N=24$ steps, and under two different strengths of the process and measurement noise.
The initial belief is $\bm{\mean}_0 = [-8, -8, 0.05, 0]$, $\cov_0 = \diag{1, 1, 0.01, 0.01}$.
To satisfy \cref{assump:PredSetRank}, we lump together every two steps, thus doubling the dimension of the input space.
We use a partition into $11 \times 5 \times 11 \times 5 = 3\,025$ regions.

\subsubsection*{Q4) Two-phase horizon}
To demonstrate the usefulness of the two-phase time horizon, we apply our method with different lengths $\bar{N}$ of the transient phase.
Recall that, at each time step, the goal and critical regions are expanded/contracted by the error bound $\varepsilon_k$ (obtained from \cref{eq:ErrorBounds}) to account for the error between the belief mean $\bm\mean_k$ and the state $\bm{x}_k$.
The results are shown in \cref{table:spacecraft} (note that \cref{tab:Benchmarks} shows the same results for $\bar{N}=3$).
We observe that increasing the length of the transient phase beyond $\bar{N} = 3$ has a negligible effect on the satisfaction probability bound $\eta^\star_{\initState}$.
At the same time, the number of \gls{iMDP} states increases linearly with the value of $\bar{N}$.
Thus, the length of the transient phase $\bar{N}$ can be used as a parameter to balance the size of the iMDPs with the resulting satisfaction guarantee obtained using \cref{thm:Correctness,thm:correctness:2phase}.

\input{Includes/Table_spacecraft}

%% file: Figures/table.tex
\begin{table*}[t!]

\caption{Overview of all benchmarks ($n$ is the state space dimension), the sizes of the iMDPs ($\bar{N}$ is the length of the transient phase), times to run \cref{alg:algorithm} (split in generating the abstraction and computing an optimal policy $\pi^\star$).
The last columns show the lower bound on the satisfaction probability guaranteed by our method, versus the empirical satisfaction in simulations.}
\label{tab:Benchmarks}

\setlength{\tabcolsep}{5pt}
\begin{tabularx}{\textwidth}{@{} lXllrrrrrr @{} } \toprule
    \multicolumn{3}{c}{\textbf{Benchmark}} & \multicolumn{3}{c}{\textbf{Abstract iMDP size}} & \multicolumn{2}{c}{\textbf{Run time of \cref{alg:algorithm}}} & \multicolumn{2}{c}{\textbf{Controller satisfaction probability}}
    \\
    \cmidrule(lr){1-3} \cmidrule(lr){4-6} \cmidrule(lr){7-8} \cmidrule(lr){9-10}
    Model & Instance & $n$ & $\bar{\finiteHorizon}$ & States & Transitions & Abstraction [s] & Compute $\pi^\star$ [s] & Guaranteed bound $(\eta^\star_{\initState})$ & Empirical ($\bar{p}_{\initState}$)
    \\ \midrule
    Pack. del. & 12x12 partition         & 2                    & 4       &    723 &       33\,210 &         0.6 &         1.3 &                   0.000 &                0.005 \\
    Pack. del. & 16x16 partition         & 2                    & 4       &   1\,283 &      101\,197 &         1.2 &         1.3 &                   0.028 &                0.631 \\
    Pack. del. & 20x20 partition         & 2                    & 4       &   2\,003 &      659\,547 &         2.3 &         3.2 &                   0.952 &                1.000 \\
    Pack. del. & 24x24 partition         & 2                    & 4       &   2\,883 &     1\,666\,695 &         4.3 &         6.9 &                   0.952 &                1.000 \\
    Pack. del. & 48x48 partition         & 2                    & 4       &  11\,523 &    90\,531\,874 &        53.7 &       307.2 &                   0.961 &                1.000 \\
    \midrule
    Spacecraft & Low noise  & 4 & 3 & 12\,103 & 3\,037\,971 & 72.3 & 9.7 & 0.929 & 1.000 \\
    Spacecraft & High noise & 4 & 3 & 12\,103 & 4\,996\,209 & 93.4 & 14.4 & 0.723 & 0.984 \\
    \midrule
    UAV 2D & Low noise ($f=0.1$) & 4 & 3 & 12\,103  & 946\,494 &        52.2 &         3.5 &                   0.983 &                1.000 \\
    UAV 2D & High noise ($f=1$) & 4 & 3 & 12\,103 & 1\,467\,391 &        65.3 &         4.5 &                   0.782 &                1.000 \\
    \midrule
    UAV 3D & Low $w_k$; low $v_k$         & 6                    & 3       & 74\,847 & 30\,817\,399 & 2,337.5 & 76.9 & 0.982 & 1.000 \\
    UAV 3D & High $w_k$; low $v_k$        & 6                    & 3       & 74\,847 & 83\,529\,922 & 2,755.7 & 182.4 & 0.731 & 0.980 \\
    UAV 3D & Low $w_k$; high $v_k$        & 6                    & 3       & 74\,847 & 40\,447\,098 & 2,349.0 & 94.8 & 0.972 & 1.000 \\
    UAV 3D & High $w_k$; high $v_k$       & 6                    & 3       & 74\,847 & 80\,431\,168 & 2,776.1 & 171.3 & 0.518 & 0.989 \\
    UAV 3D &  High $w_k$; high $v_k$ in Z dir. & 6               & 3       & 74\,847 & 82\,828\,351 & 2,734.1 & 177.3 & 0.628 & 0.994 \\
    UAV 3D &  High $w_k$; high $v_k$ in Y dir. & 6               & 3       & 74\,847 & 82\,732\,697 & 2,763.5 & 172.7 & 0.653 & 0.988 \\
    \bottomrule
\end{tabularx}

\end{table*}

%% file: Includes/Table_spacecraft.tex
\begin{table}[t!]
\centering

\caption{Size of the abstract \gls{iMDP}, abstraction time, and satisfaction probability $p^\star_{\initState}$ as a function of the length of the transient phase $\bar{N}$ of the two-phase horizon.}
\setlength{\tabcolsep}{4.5pt}
\begin{tabular}{l|lrrrrr}
    \toprule
    \textbf{Noise} & \textbf{Tran. phase $\bar{N}$} &
    \textbf{1} & \textbf{2} & \textbf{3} & \textbf{4} & \textbf{5} \\
    \midrule
    \multirow{3}{*}{Low} & \gls{iMDP} states $|S|$ & 
    $6.1\text{K}$ & $9.1\text{K}$ & $12.1\text{K}$ & $15.1\text{K}$ & $18.1\text{K}$
    \\
    & Lower bound $\eta^\star_{\initState}$ &
    0.207 & 0.649 & 0.929 & 0.929 & 0.929
    \\
    & Abstraction time [s] &
    56.8 & 60.8 & 72.3 & 86.6 & 92.7
    \\
    \midrule
    \multirow{3}{*}{High} & \gls{iMDP} states $|S|$ & 
    $6.1\text{K}$ & $9.1\text{K}$ & $12.1\text{K}$ & $15.1\text{K}$ & $18.1\text{K}$
    \\
    & Lower bound $\eta^\star_{\initState}$ &
    0.115 & 0.343 & 0.723 & 0.728 & 0.728
    \\
    & Abstraction time [s] &
    77.1 & 81.9 & 93.4 & 103.9 & 121.8
    \\
\bottomrule
\end{tabular}

\label{table:spacecraft}

\end{table}

%% file: 8-Conclusions.tex
\subsection{Comparison with existing methods}
Let us explain again why solving \cref{prob:Problem} is infeasible with the alternative methods discussed in the related work in \cref{sec:related_work}.
The key characteristic of our method that enables us to solve \cref{prob:Problem} is that our method yields a \emph{formal lower bound guarantee} (namely, the bound $\eta^\star_{\initState}$) on the probability to satisfy a property.
That is, when the obtained feedback controller is applied to the concrete \gls{LTI} system, the reach-avoid property is satisfied with \emph{at least} a probability of $\eta^\star_{\initState}$.

\change{In \cref{subsec:experiments:UAV}, we have demonstrated that the popular RRBT cannot provide formal guarantees on property satisfaction under control input constraints.
Moreover, recall from \cref{sec:related_work} that methods such as FIRM~\cite{DBLP:conf/iros/Agha-mohammadiCA11} and SLAP~\cite{DBLP:journals/trob/Agha-mohammadiA18} rely on maximum likelihood estimates (MLEs), leading to \emph{approximations} of the satisfaction probability with \emph{statistical errors}.
It has been shown empirically by~\cite{DBLP:conf/aaai/BadingsA00PS22} that using MLEs does not lead to sound bounds on the satisfaction probability.
Thus, these methods cannot solve \cref{prob:Problem}, as this problem requires a \emph{hard lower bound} on the satisfaction probability.}

Also recall from \cref{sec:related_work} that, while control barrier functions (CBFs) can provide formal guarantees on the satisfaction of reach-avoid properties via, e.g., optimization, the tractability of these methods strongly depends on the convexity of the problem~\cite{clark2021control,DBLP:journals/tac/AmesXGT17,DBLP:journals/tac/PrajnaJP07}.
Thus, most practical applications (including those in the references above) consider problems with convex safe sets, which is not the case in our benchmarks (as clearly shown by \cref{fig:UAV_3D}).
By contrast, the complexity of our method is independent of the convexity of the safe set.

These advantages of our approach do come at a significant computational cost.
The size of abstractions tends to scale exponentially with the partition resolution and dimension of the state space, commonly called the \emph{curse of dimensionality}.
Moreover, since our formal guarantees rely on the optimality of the Kalman filter, our method is limited to linear systems.
Finally, we have only considered reach-avoid properties in this paper.
In \cref{sec:Conclusions}, we describe several directions for future work that aim to mitigate some of these limitations.

\section{Conclusion}
\label{sec:Conclusions}

We have provided a correct-by-construction controller synthesis scheme for \gls{LTI} systems with Gaussian noise based on Kalman filtering.
This approach allows us to soundly abstract a continuous-state system into a finite-state \gls{MDP} with intervals of transition probabilities.
The numerical experiments show that our approach synthesizes feedback controllers that satisfy reach-avoid specifications across several domains.

One fundamental limitation of our approach is the limited scalability.
To address the computational limitations, we believe that hybrid schemes that combine, for example, sample-based methods (to search for candidate solutions) with abstraction (to verify these candidate solutions) are of particular interest.
In future research, we thus wish to integrate our abstraction-based scheme with a sample-based algorithm in such a manner.
We also wish to explore adaptive schemes for discretizing the state space~\cite{DBLP:journals/siamads/SoudjaniA13}, so that we refine a coarse initial discretization only when we benefit from it.

Finally, our focus on reach-avoid properties is without loss of generality, and our scheme can directly be applied to any specification for which \gls{iMDP} model checking is possible. 
For details, we refer to~\cite{DBLP:journals/corr/Rickard}, which uses an abstraction scheme similar to ours for general \gls{PCTL} model checking.
Extensions beyond \gls{PCTL} to, e.g., \gls{LTL} lead to questions regarding the semantics of transition probability intervals.
For example, is the uncertainty in the probability distributions \emph{static} (i.e., the same probability distribution is chosen in each encounter of the same state-action pair) or \emph{dynamic} (i.e., a different probability distribution can be chosen in each encounter)~\cite{DBLP:journals/mor/Iyengar05}?
Further research is necessary to answer such questions.